%% file: PAMI_2017_DR_GAN.tex
\documentclass[10pt,journal,compsoc]{IEEEtran}
\ifCLASSOPTIONcompsoc
  \usepackage[nocompress]{cite}
\else
  \usepackage{cite}
\fi
\usepackage{amsmath}
\usepackage{epsfig}
\usepackage{graphicx}
\usepackage{array}
\usepackage{multirow}
\usepackage{amssymb}
\usepackage{extarrows}
\usepackage{booktabs}       
\usepackage{diagbox}
\usepackage{comment}

\usepackage{times}
\usepackage{wrapfig}
\usepackage{caption}
\usepackage{subcaption}
\usepackage{verbatim}
\usepackage{color}
\usepackage{cite}

\usepackage{forloop}

\usepackage{url}

\input{newcommands.tex}

\begin{document}
%
\title{Representation Learning by Rotating Your Faces}

\author{Luan~Tran,
        Xi~Yin, 
        and~Xiaoming~Liu,~\IEEEmembership{Member,~IEEE}
\IEEEcompsocitemizethanks{\IEEEcompsocthanksitem L. Tran, X. Yin and X. Liu are with the Department of Computer Science and Engineering, Michigan State University. E-mail: tranluan@msu.edu, yinxi1@msu.edu, liuxm@cse.msu.edu 
}%
}

\markboth{IEEE TRANSACTIONS ON PATTERN ANALYSIS AND MACHINE INTELLIGENCE}%
{Representation Learning by Rotating Your Faces}
%



\IEEEtitleabstractindextext{%
\begin{abstract}
The large pose discrepancy between two face images is one of the fundamental challenges in automatic face recognition.
Conventional approaches to pose-invariant face recognition either perform face frontalization on, or learn a pose-invariant representation from, a non-frontal face image.
We argue that it is more desirable to perform both tasks jointly to allow them to leverage each other. 
To this end, this paper proposes a Disentangled Representation learning-Generative Adversarial Network (DR-GAN) with three distinct novelties.
First, the encoder-decoder structure of the generator enables DR-GAN to learn a representation that is both generative and discriminative, which can be used for face image synthesis and pose-invariant face recognition.
Second, this representation is explicitly disentangled from other face variations such as pose, through the pose code provided to the decoder and pose estimation in the discriminator.
Third, DR-GAN can take one or multiple images as the input, and generate one unified identity representation along with an arbitrary number of synthetic face images. 
Extensive quantitative and qualitative evaluation on a number of controlled and in-the-wild databases demonstrate the superiority of DR-GAN over the state of the art in both learning representations  and rotating large-pose face images. 
\end{abstract}

}

\maketitle

\IEEEdisplaynontitleabstractindextext

%
\IEEEpeerreviewmaketitle

\input{PAMI_2017_DR_GAN_intro.tex}
\input{PAMI_2017_DR_GAN_prior.tex}

\input{PAMI_2017_DR_GAN_alg.tex}

\input{PAMI_2017_DR_GAN_exp.tex}
\input{PAMI_2017_DR_GAN_con.tex}


%

%

\ifCLASSOPTIONcompsoc
\else
\fi


\ifCLASSOPTIONcaptionsoff
  \newpage
\fi



\bibliographystyle{IEEEtran}
\bibliography{dr_gan}
%

%

%
\vspace{-7mm}
\input{bio/LuanTran.tex}
\vspace{-7mm}
\input{bio/XiYin.tex}
\vspace{-7mm}
\input{bio/XiaomingLiu.tex}




\end{document}

%% file: newcommands.tex
\newcommand\T{{\hspace{-1pt}\intercal}}

\newcommand\etal{\textit{et~al.}}

\newcommand{\DrGanAM}{DR-GAN$_{\text{AM}}$}

\newcommand{\Paragraph}[1]{\vspace{-0mm} \noindent \textbf{#1.} \hspace{0mm}}

\newcommand{\Section}[1]{\vspace{0mm} \section{#1} \vspace{0mm}}
\newcommand{\SubSection}[1]{\vspace{0mm} \subsection{#1} \vspace{0mm}}
\newcommand{\SubSubSection}[1]{\vspace{0mm} \subsubsection{#1} \vspace{0mm}}

\newcounter{ct}

\def\eqnvspace{{\vspace{0mm}}}
\def\figvspace{{\vspace{-2mm}}}

\def\interpolation_collumsize{0.064\textwidth}

\def\feature_block_size{0.04\textwidth}

%% file: PAMI_2017_DR_GAN_intro.tex
\IEEEraisesectionheading{\section{Introduction}\label{sec:introduction}}

\IEEEPARstart{F}{ace} recognition is one of the most widely studied topics in computer vision due to its wide application in law enforcement, biometrics, marketing, and etc.
Recently, great progress has been achieved in face recognition with deep learning-based methods~\cite{taigman2014deepface, parkhi2015deep, schroff2015facenet}.
For example, surpassing human performance is reported by Schroff et al.~\cite{schroff2015facenet} on Labeled
Faces in the Wild (LFW) database. 
However, one of the shortcomings of the LFW database is that it does not offer a high degree of pose variation --- the variance that has been shown to be a major challenge in face recognition. 
Up to now, the key ability of Pose-Invariant Face Recognition (PIFR) desired by real-world applications is far from solved~\cite{liu2005pose, liu2006optimal, chai2007locally, abiantun2014sparse, ding2016comprehensive}.
A recent study~\cite{sengupta2016frontal} observes a significant drop, over $10\%$, in performance of most algorithms from frontal-frontal to frontal-profile face verification, while human performance only degrades slightly.
This indicates that the pose variation remains to be a significant challenge in face recognition and warrants future study.

In PIFR, the facial appearance change caused by pose variation often significantly surpasses the intrinsic appearance differences between individuals. 
To overcome these challenges, a wide variety of approaches have been proposed, which can be grouped into two categories.
First, some work employ {\it face frontalization} on the input image to synthesize a frontal-view face, where traditional face recognition algorithms are applicable~\cite{hassner2015effective, zhu2015high}, or an identity representation can be obtained via modeling the face frontalization/rotation process~\cite{kan2014stacked,zhu2014multi,yim2015rotating}.
The ability to generate a realistic identity-preserved frontal face is also beneficial for law enforcement practitioners to identify suspects.
Second, other work focus on {\it learning discriminative representations} directly from the non-frontal faces through either one joint model~\cite{parkhi2015deep,schroff2015facenet} or multiple pose-specific models~\cite{masi2016pose,ding2015robust}.
In contrast, we propose a novel framework to take the best of both worlds --- {\it simultaneously learn pose-invariant identity representation} and {\it synthesize faces with arbitrary poses}, where face rotation is both a facilitator and a by-product for representation learning.

\begin{figure}[t!]
\centering
\includegraphics[trim=0 80 165 66, clip, width=0.8\linewidth]{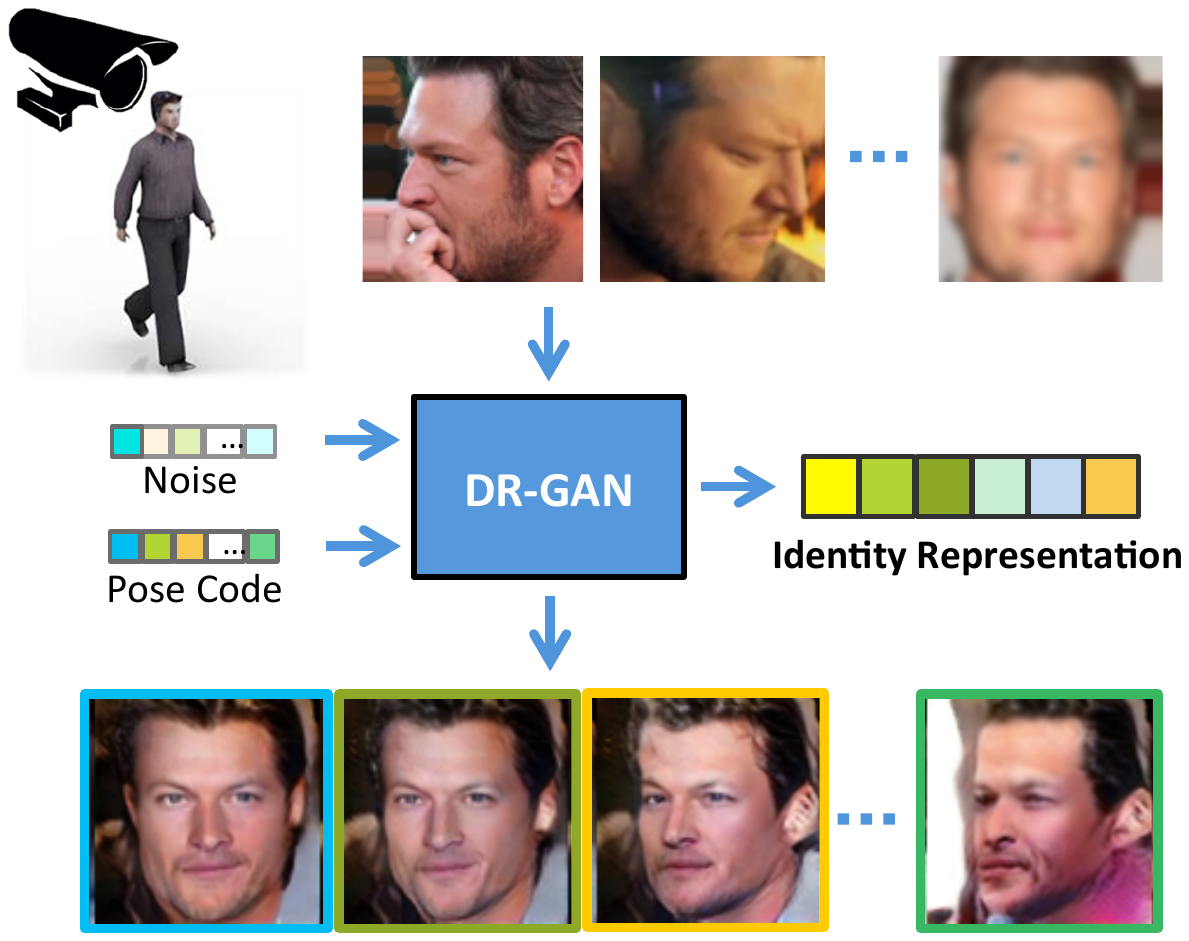}
\vspace{-1mm}
\caption{\small Given one or multiple in-the-wild face images as the input, DR-GAN can produce a unified identity representation, by virtually rotating the face to arbitrary poses. The learnt representation is both {\it discriminative} and {\it generative}, i.e., the representation is able to demonstrate superior PIFR performance, and synthesize identity-preserved faces at target poses specified by the pose code.}
\label{fig:concept}
\figvspace 
\end{figure}

As shown in Fig.~\ref{fig:concept}, we propose Disentangled Representation learning-Generative Adversarial Network (DR-GAN) for PIFR.
Generative Adversarial Networks~(GANs)~\cite{goodfellow2014generative} can generate samples following a data distribution through a two-player game between a generator $G$ and a discriminator $D$.
Despite many recent promising developments~\cite{mirza2014conditional, denton2015deep, radford2015unsupervised, chen2016infogan, berthelot2017began}, image synthesis remains to be the main objective of GAN.
To the best of our knowledge, this is the first work that utilizes the generator in GAN for representation learning.
To achieve this, we conduct $G$ with an encoder-decoder structure (Fig.~\ref{fig:gans} (d)) to learn a disentangled representation for PIFR.
The input to the encoder $G_{enc}$ is a face image of any pose, the output of the decoder $G_{dec}$ is a synthetic face at a target pose, and the learnt representation bridges $G_{enc}$ and $G_{dec}$. 
While $G$ serves as a face rotator, $D$ is trained to not only distinguish real vs.~synthetic (or fake) images, but also predict the identity and pose of a face.
With the additional classifications, $D$ strives for the rotated face to have the same identity as the input real face, which has two effects on $G$: 
1) The rotated face looks more like the input subject in terms of identity. 
2) The learnt representation is more {\it inclusive} or {\it generative} for synthesizing an identity-preserved face. 

In conventional GANs, $G$ takes a random noise vector to synthesize an image.
In contrast, our $G$ takes a face image, a pose code $\bf{c}$, and a random noise vector $\bf{z}$ as the input, with the objective of generating a face of the same identity with the target pose that can fool $D$.
Specifically, $G_{enc}$ learns a mapping from the input image to a feature representation. 
The representation is then concatenated with the pose code and the noise vector to feed to $G_{dec}$ for face rotation. 
The noise models facial appearance variations other than identity or pose. 
Note that it is a crucial architecture design to concatenate one representation with {\it varying} randomly generated pose codes and noise vectors.
This enables DR-GAN to learn a {\it disentangled} identity representation that is {\it exclusive} or {\it invariant} to pose and other variations, which is the holy grail for PIFR when achievable. 

Most existing face recognition algorithms only takes one image for testing. 
In practice, there are many scenarios when an image collection of the same individual is available~\cite{klare2015pushing}. 
In this case, prior work fuse results either in the feature level~\cite{chen2016unconstrained} or the distance-metric level~\cite{wang2016face,masi2016we}. 
Differently, our fusion is conducted within a unified framework. 
Given multiple images as the input, $G_{enc}$ operates on each image, and produces an identity representation and a coefficient, which is an indicator of the quality of that input image. 
Using the dynamically learned coefficients, the representations of all input images are linearly combined as one representation. 
During testing, $G_{enc}$ takes any number of images and generates a single identity representation, which is used by $G_{dec}$ for face synthesis along with the pose code. 

Our generator is essential to both representation learning and image synthesis. 
We propose two techniques to further improve $G_{enc}$ and $G_{dec}$ respectively. 
First, we have observed that our $G_{enc}$ can always outperform $D$ in representation learning for PIFR. 
Therefore, we propose to replace the identity classification part of $D$ with the latest $G_{enc}$ during training so that a superior $D$ can push $G_{enc}$ to further improve itself. 
Second, since our $G_{dec}$ learns a mapping from the feature space to the image space, we propose to improve the learning of $G_{dec}$ by regularizing the average representation of two representations from different subjects to be a valid face, assuming a convex space of face identities. 
These two techniques are shown to be effective in improving the generalization ability of DR-GAN. 

A preliminary version of this work was published in 2017 IEEE Conference on Computer Vision and Pattern Recognition~\cite{tran2017disentangled}. 
We extend it in numerous ways: 1) Instead of having an extra dimension of the fake class in the identity classification task of the discriminator, we split it into two tasks: real/fake and identity classification. 
2) We propose two techniques to improve model generalization during training. 
3) We conduct all experiments using the new models with color image input, and add numerous experiments to reveal how DR-GAN works including the disentangled representation, the coefficients analysis, etc. 

In summary, this paper makes the following contributions.

\begin{itemize}
 \item  We propose DR-GAN via an encoder-decoder structured generator that can frontalize or rotate a face with an arbitrary pose, even the extreme profile.

 \item  Our learnt representation is explicitly disentangled from the pose variation via the pose code in the generator and the pose estimation in the discriminator. 
Similar disentanglement is conducted for other variations, e.g., illumination.

 \item  We propose a novel scheme to adaptively fuse multiple faces to a single representation based on the learnt coefficients, which empirically shows to be a good indicator of the face image quality. 

 \item  We propose two techniques to improve the generalization ability of our generator via model switch and representation interpolation. 

 \item  We achieve state-of-the-art face frontalization and face recognition performance on multiple benchmark datasets, including Multi-PIE~\cite{gross2010multi}, CFP~\cite{sengupta2016frontal}, and IJB-A~\cite{klare2015pushing}.
\end{itemize}

%% file: PAMI_2017_DR_GAN_prior.tex
\section{Prior Work}

\Paragraph{Generative Adversarial Network (GAN)}
Goodfellow \etal~\cite{goodfellow2014generative} introduce GAN to learn generative models via an adversarial process. 
With a minimax two-player game, the generator and discriminator can both improve themselves. 
GAN has been used for image synthesis~\cite{denton2015deep,reed2016generative}, image super resolution~\cite{yu2016ultra}, and etc. 
More recent work focus on incorporating constraints to $\bf{z}$ or leveraging side information for better synthesis. 
E.g., Mirza and Osindero~\cite{mirza2014conditional} feed class labels to both $G$ and $D$ to generate images conditioned on class labels. 
In~\cite{salimans2016improved} and~\cite{odena2016semi}, GAN is generalized to learn a discriminative classifier where $D$ is trained to not only distinguish between real vs.~fake, but also classify the images. 
In InfoGAN~\cite{chen2016infogan}, $G$ applies information regularization to the optimization by using the additional latent code. In contrast, this paper proposes a novel DR-GAN aiming for face {\it representation learning}, 
which is achieved via modeling the face rotation process. In Sec.~\ref{sec:gans}, we will provide in-depth discussion on our difference to most relevant work in GANs.

One crucial issue with GANs is the difficulty for quantitative evaluation. 
Previous work either perform human study to evaluate the quality of synthetic images~\cite{denton2015deep} or use the features in the discriminator for image classification~\cite{radford2015unsupervised}.
In contrast, we innovatively construct the generator for representation learning, which can be quantitatively evaluated for PIFR.

\begin{figure*}[t!]
\centering
\includegraphics[trim=0 0 10 0,clip, width=0.85\linewidth]{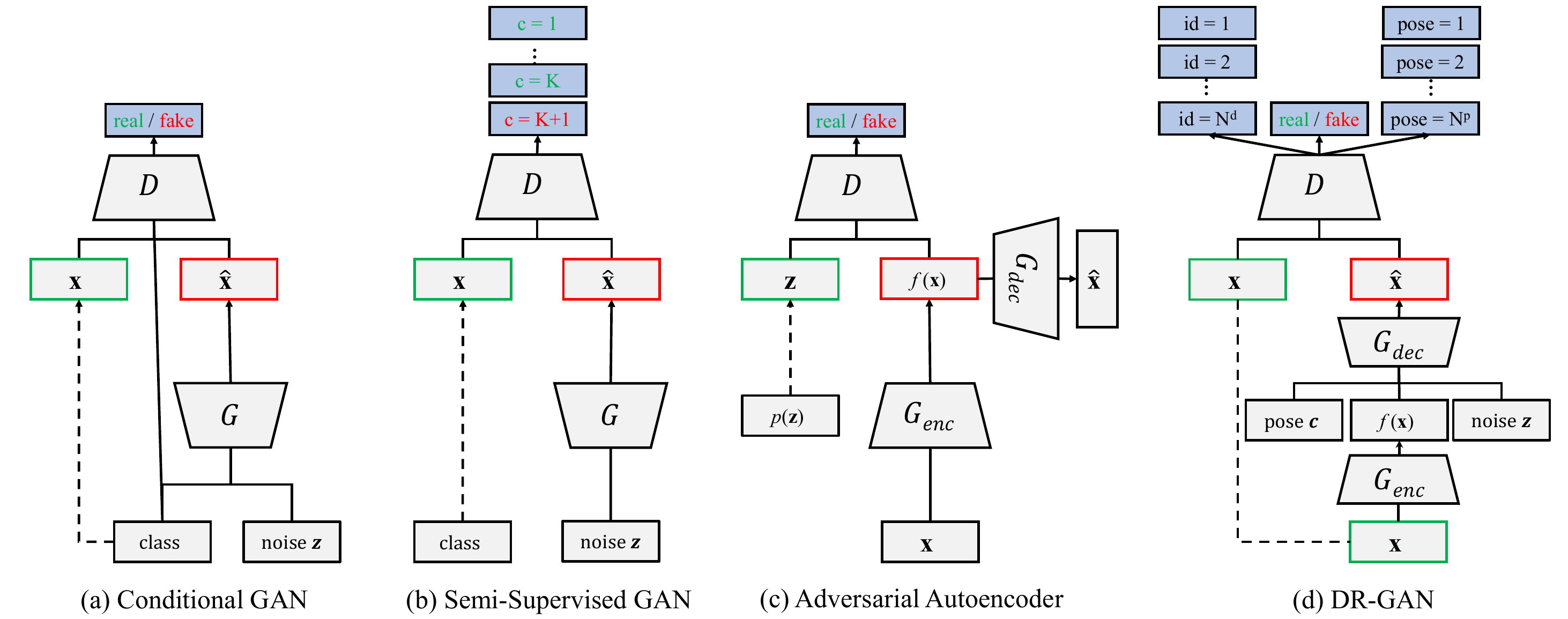}
\caption{\small Comparison of previous GAN architectures and our proposed DR-GAN.}
\label{fig:gans}
\figvspace
\end{figure*}

\Paragraph{Face Frontalization}
Generating a frontal face from a profile face is very challenging due to self-occlusion. 
Prior methods in face frontalization can be classified into three categories: $3$D-based methods~\cite{zhu2015high, hassner2015effective,li2012morphable}, statistical methods~\cite{sagonas2015robust}, and deep learning methods~\cite{zhu2014multi, yang2015weakly, yim2015rotating,kan2014stacked,zhang2013random}. 
E.g., Hassner \etal~\cite{hassner2015effective} use a mean $3$D face model to generate a frontal face for any subject.
A personalized face model could be used but accurate $3$D face reconstruction remains a challenge~\cite{roth2017pami,joint-face-alignment-and-3d-face-reconstruction,tran2018nonlinear,tran2018on}.
In~\cite{sagonas2015robust}, a statistical model is used for joint frontalization and landmark localization by solving a constrained low-rank minimization problem. 
For deep learning methods, Kan \etal~\cite{kan2014stacked} propose SPAE to progressively rotate a non-frontal face to a frontal one via auto-encoders.  
Yang \etal~\cite{yang2015weakly} apply the recurrent action unit to a group of hidden units to incrementally rotate faces in fixed yaw angles. 

All prior work frontalize only near frontal in-the-wild faces~\cite{hassner2015effective,zhu2015high} or large-pose controlled faces~\cite{yim2015rotating,zhu2014multi}.
In contrast, we can synthesize arbitrary-pose faces from a large-pose in-the-wild face. 
We use the {\it adversarial loss} to improve the quality of the synthetic images and identity classification in the discriminator to preserve identity. 

\Paragraph{Representation Learning} 
Designing the appropriate objectives for learning a good representation is an open question~\cite{bengio2013representation}.
The work in~\cite{huang2007unsupervised} is among the first to use an encoder-decoder structure for representation learning, which, however, is not explicitly disentangled. 
DR-GAN is similar to DC-IGN~\cite{kulkarni2015deep} --- a variational autoencoder-based method to disentangled representation learning. 
However, DC-IGN achieves disentanglement by providing batch training samples with one attribute being fixed, which may not be applicable to unstructured in-the-wild data. 

Prior work also explore joint representation learning and face rotation for PIFR where~\cite{zhu2014multi,yim2015rotating} are most relevant to our work. 
In~\cite{zhu2014multi}, Multi-View Perceptron~\cite{zhu2014multi} is used to untangle the identity and view representations by processing them with different neurons and maximizing the data log-likelihood. 
Yim \etal~\cite{yim2015rotating} use a multi-task CNN to rotate a face with any pose and illumination to a target pose, and the $L2$ loss-based reconstruction of the input is the second task.
Both work focus on image synthesis and the identity representation is a by-product during the network learning. 
In contrast, DR-GAN focuses on representation learning, of which face rotation is both a facilitator and a by-product.
We differ to~\cite{zhu2014multi,yim2015rotating} in four aspects. 
First, we explicitly disentangle the identity representation from pose variations by pose codes. 
Second, we employ the adversarial loss for high-quality synthesis, which drives better representation learning. 
Third, none of them applies to in-the-wild faces as we do. 
Finally, our ability to learn the representation from multiple unconstrained images has not been observed in prior work.

\Paragraph{Face Image Quality Estimation}
Low image quality is known to be a challenge for vision tasks~\cite{face-model-fitting-on-low-resolution-images, fsrnet-end-to-end-learning-face-super-resolution-with-facial-priors}.
Image quality estimation is important for biometric recognition systems~\cite{bharadwaj2014biometric,grother2007performance, tong2010improving}.
Numerous methods have been proposed to measure the image quality of different biometric modalities including face~\cite{abaza2014design,abdel2007application,ozay2009improving}, iris~\cite{chen2006localized,krichen2007new}, fingerprint~\cite{tabassi2005novel,teixeira2016new}, and gait~\cite{muramatsu2016view,matovski2012including}. 
In the scenario of face recognition, 
an effective algorithm for face image quality estimation can help to either (i) reduce the number of poor images acquired during enrollment, or (ii) improve feature fusion during testing. 
Both cases can improve the face recognition performance. 
Abaza~\etal~\cite{abaza2014design} evaluate multiple quality factors such as contrast, brightness, sharpness, focus and illumination as a face image quality index for face recognition. 
However, they did not consider pose variance, which is a major challenge in face recognition. 
Ozay~\etal~\cite{ozay2009improving} employ a Bayesian network to model the relationships between predefined quality related image features and face recognition, which is show to boost the performance significantly. 
The authors in~\cite{wong2011patch} propose a patch-based face image quality estimation method, which takes into account of geometric alignment, pose, sharpness, and shadows. 

In this work, we employ quality estimation in a unified GAN framework that considers all factors of image quality presented in the dataset, with {\it no} direct supervision.
For each input image, DR-GAN can generate a coefficient that indicates the quality of the input image. 
The representations from multiple images of the same subject are fused based on the learnt coefficients to generate one unified representation. 
We will show that the learnt coefficients are correlated to the image quality, i.e., a measurement of how good it can be used for face recognition.

%% file: PAMI_2017_DR_GAN_alg.tex
\section{The Proposed DR-GAN Model}
\label{sec:method}
Our proposed DR-GAN has two variations: the basic model can take one image per subject for training, termed {\it single-image DR-GAN}, and the extended model can leverage multiple images per subject for both training and testing, termed {\it multi-image DR-GAN}. 
We start by introducing the original GAN, followed by two DR-GAN variations, and the proposed techniques to improve the generalization of our generator. 
Finally, we will compare our DR-GAN with previous GAN variations in detail. 

\input{PAMI_2017_DR_GAN_alg_3.1.tex}

\input{PAMI_2017_DR_GAN_alg_3.2.tex}

\input{PAMI_2017_DR_GAN_alg_3.3.tex}

\input{PAMI_2017_DR_GAN_alg_3.4.tex}

\input{PAMI_2017_DR_GAN_alg_3.5.tex}

%% file: PAMI_2017_DR_GAN_alg_3.1.tex

\SubSection{Generative Adversarial Network} 
Generative Adversarial Network consists of a generator $G$ and a discriminator $D$ that compete in a two-player minimax game.
The discriminator $D$ tries to distinguish between a real image $\bf{x}$ and a synthetic image $G(\bf{z})$.
The generator $G$ tries to synthesize realistic-looking images from a random noise vector $\bf{z}$ that can fool $D$, i.e., $G(\bf{z})$ being classified as a real image. 
Concretely, $D$ and $G$ play the game with the following loss function:
\eqnvspace
\begin{align}
\min_{G} \max_{D} \mathcal{L}_{gan} = {} & \mathbb{E}_{\mathbf{x} \sim p_{d}(\mathbf{x})} [\log D(\mathbf{x})] + \nonumber \\ 
& \mathbb{E}_{\mathbf{z} \sim p_{z}(\mathbf{z})} [\log(1- D(G(\mathbf{z}))) ].
\eqnvspace
\end{align}

It is proved in~\cite{goodfellow2014generative} that this minimax game has a global optimum when the distribution $p_{g}$ of the synthetic samples and the distribution $p_{d}$ of the real samples are the same. 
Under mild conditions (e.g., $G$ and $D$ have enough capacity), $p_{g}$ converges to $p_{d}$. 
In the beginning of training, the samples generated from $G$ are extremely poor and are rejected by $D$ with high confidences.
In practice, it is better for $G$ to maximize $\log(D(G(\mathbf{z})))$ instead of minimizing $\log \left( 1 - D(G(\mathbf{z})) \right)$~\cite{goodfellow2014generative}. This objective results in the
same fixed point of the dynamics of $G$ and $D$ but provides much stronger gradients early in learning.
As a result, $G$ and $D$ are trained to alternatively optimize the following objectives:
\vspace{-1mm}
\begin{align}
\max_{D} \mathcal{L}_{gan}^{D} = {} & \mathbb{E}_{\mathbf{x} \sim p_{d}(\mathbf{x})} [\log D({\bf{x}})] + \nonumber \\ 
 & \mathbb{E}_{\mathbf{z} \sim p_{z}(\mathbf{z})} [\log(1- D(G(\mathbf{z}))) ], \\
\max_{G} \mathcal{L}_{gan}^{G} = {} & \mathbb{E}_{\mathbf{z} \sim p_{z}(\mathbf{z})} [\log(D(G(\mathbf{z})) ].
\eqnvspace\vspace{-5mm}
\end{align}


%% file: PAMI_2017_DR_GAN_alg_3.2.tex
\vspace{-2mm}
\SubSection{Single-Image DR-GAN} 
\label{sec:single_image_model}
Our single-image DR-GAN has two distinctive novelties compared to prior GANs.
First, it learns an identity representation for a face image by using an encoder-decoder structured generator, where the representation is the encoder's output and the decoder's input.
Since the representation is the input to the decoder to synthesize various faces of the same subject, i.e., virtually rotating his/her face, it is a {\it generative} representation.

Second, the appearance of a face is determined by not only the identity, but also the numerous distractive variations, such as pose, illumination, expression.
Thus, the identity representation learned by the encoder would inevitably include the distractive side variations.
E.g., the encoder would generate {\it different} identity representations for two faces of the same subject with $0^{\circ}$ and $90^{\circ}$ yaw angles.
To remedy this, in addition to the class labels similar to semi-supervised GAN~\cite{salimans2016improved}, we employ side information such as pose and illumination to explicitly disentangle these variations, which in turn helps to learn a {\it discriminative} representation.

\SubSubSection{Problem Formulation}
Given a face image $\mathbf{x}$ with label $\mathbf{y} = \{ y^{d}, y^{p}\}$, where $y^d$ represents the label for identity and $y^p$ for pose, the objectives of our learning problem are twofold: 1) to learn a pose-invariant identity representation for PIFR, and 2) to synthesize a face image $\hat{\mathbf{x}}$ with the {\it same} identity $y^{d}$ but at a {\it different} pose specified by a pose code $\mathbf{c}$. 
Our approach is to train a DR-GAN conditioned on the original image $\mathbf{x}$ and the pose code $\mathbf{c}$ with its architecture illustrated in Fig.~\ref{fig:gans} (d).

Different from the discriminator in conventional GAN, our $D$ is a multi-task CNN consisting of three components: $D = [D^r, D^d, D^p]$. 
$D^r\in \mathbb{R}^{1}$ is for real/fake image classification. 
$D^d\in \mathbb{R}^{N^d}$ is for identity classification with $N^d$ as the total number of subjects in the training set.
$D^p\in \mathbb{R}^{N^p}$ is for pose classification with $N^p$ as the total number of discrete poses.
Note that, in our preliminary work~\cite{tran2017disentangled}, $D^r$ is implemented as an additional $N^d+1^{th}$ element of $D^d$, which has the problem of unbalanced training data for each dimension in $D^d$, i.e., the number of synthetic images ($N^d+1^{th}$ dimension) equals to the summation of all images in the real classes (the first $N^d$ dimensions). 
This version fixes this problem and is referred as ``split" in Tab.~\ref{tab:ijb-a}.
Given a face image $\mathbf{x}$,  $D$ aims to classify it as the real image class, and estimate its identity and pose; while given a synthetic face image from the generator $\hat{\mathbf{x}} = G(\mathbf{x}, \mathbf{c}, \mathbf{z})$, $D$ attempts to classify $\hat{\mathbf{x}}$ as fake, using the following objectives:
\begin{align}
 \mathcal{L}_{gan}^{D}  = {}& \mathbb{E}_{\mathbf{x} ,\mathbf{y} \sim p_{d}(\mathbf{x},\mathbf{y})} [\log D^r({\bf{x}})] + \nonumber \\ 
 {}& \mathbb{E}_{\substack{\mathbf{x} ,\mathbf{y} \sim p_{d}(\mathbf{x},\mathbf{y}), \\ \mathbf{z} \sim p_{z}(\mathbf{z}) , \mathbf{c} \sim p_{c}(\mathbf{c}) }} [\log(1- D^r(G(\mathbf{x}, \mathbf{c}, \mathbf{z}))) ], \\
\mathcal{L}_{id}^{D}  = {} & \mathbb{E}_{\mathbf{x} ,\mathbf{y} \sim p_{d}(\mathbf{x},\mathbf{y})} [\log D^d_{y^d}({\bf{x}})], \\
\mathcal{L}_{pos}^{D}   = {} & \mathbb{E}_{\mathbf{x} ,\mathbf{y} \sim p_{d}(\mathbf{x},\mathbf{y})} [\log D^p_{y^p}({\bf{x}})],
\end{align}



\noindent where $D^d_i$ and $D^{p}_i$ are the $i$th element in $D^d$ and $D^p$. 
For clarity, we will eliminate all subscripts for expected value notations, as all random variables are sampled from their respected distributions $(\mathbf{x} ,\mathbf{y} \sim p_{d}(\mathbf{x},\mathbf{y}), \mathbf{z} \sim p_{z}(\mathbf{z}) , \mathbf{c} \sim p_{c}(\mathbf{c}))$. 
The final objective for training $D$ is the weighted average of all objectives:
\begin{equation}
  \max_{D} \mathcal{L}^{D} = \lambda_{g} \mathcal{L}_{gan}^{D} + \lambda_{d} \mathcal{L}_{id}^{D} +\lambda_{p} \mathcal{L}_{pos}^{D},
\label{eqn:objD}
\end{equation}
where we set $\lambda_{g} = \lambda_{d} = \lambda_{p} = 1$.

Meanwhile, $G$ consists of an encoder $G_{enc}$ and a decoder $G_{dec}$. 
$G_{enc}$ aims to learn an identity representation $f({\bf{x}}) = G_{enc}({\bf{x}})$ from a face image $\mathbf{x}$. 
$G_{dec}$ aims to synthesize a face image $\hat{\mathbf{x}} = G_{dec}(f({\bf{x}}), \mathbf{c}, \mathbf{z})$ with identity $y^{d}$ and a target pose specified by $\mathbf{c}$, and  $\mathbf{z}\in \mathbb{R}^{N^z}$ is the noise  modeling other variations besides identity or pose. 
The pose code ${\bf{c}}\in \mathbb{R}^{N^p}$ is a one-hot vector with the target pose $y^t$ being $1$. 
The goal of $G$ is to fool $D$ to classify $\hat{\mathbf{x}}$ to the identity of input $\mathbf{x}$ and the target pose with the following objectives:
\begin{align}
 \mathcal{L}_{gan}^{G} = {} & \mathbb{E}[\log D^r({G(\mathbf{x}, \mathbf{c}, \mathbf{z})})],  \\ 
 \mathcal{L}_{id}^{G} = {} & \mathbb{E}[\log D^d_{y^d}({G(\mathbf{x}, \mathbf{c}, \mathbf{z})})], \\
\mathcal{L}_{pos}^{G} = {} &  \mathbb{E}[\log D^p_{y^t}({G(\mathbf{x}, \mathbf{c}, \mathbf{z})})].
\end{align}


\vspace{1mm}

Similarly, the final objective for training the discriminator $G$ is the weighted average of each objective:
\begin{equation}
  \max_{G} \mathcal{L}^{G} = \mu_{g} \mathcal{L}_{gan}^{G} + \mu_{d} \mathcal{L}_{id}^{G} +\mu_{p} \mathcal{L}_{pos}^{G},
\label{eqn:objG}
\end{equation}
where we set $\mu_{g} = \mu_{d} = \mu_{p} = 1$.


$G$ and $D$ improves each other during the alternative training process.
With $D$ being more powerful in distinguishing real vs.~fake images and classifying poses, $G$ strives for synthesizing an identity-preserved face with the target pose to compete with $D$. 
We benefit from this process in three aspects.
First, the learnt representation $f({\bf{x}})$ will preserve more discriminative identity information. 
Second, the pose classification in $D$ guides the pose of the rotated face to be more accurate. 
Third, with a separate pose code as input to $G_{dec}$, $G_{enc}$ is trained to disentangle the pose variation from $f({\bf{x}})$, i.e., $f({\bf{x}})$ should encode as {\it much} identity information as possible, but as {\it little} pose information as possible.
Therefore, $f({\bf{x}})$ is not only generative for image synthesis, but also discriminative for PIFR. 

\SubSubSection{Network Structure}

\begin{table}[t!]
\caption{\small The structures of $G_{enc}$, $G_{dec}$ and $D$ networks in single-image and  multi-image DR-GAN. {\color{blue}{Blue}} texts represent extra elements to learn the coefficient $\omega$  in the $G_{enc}$ of multi-image DR-GAN.} 
\label{tab:network}
\figvspace 
\begin{center}
\small
\resizebox{0.99\linewidth}{!}{
\setlength{\tabcolsep}{3pt}
\begin{tabular}{ @{}ccccccccc@{} }
\toprule
\multicolumn{3}{c}{$G_{enc}$ and $D$} & \hspace{2mm} 
& \multicolumn{3}{c}{$G_{dec}$} \\
\cmidrule(r){1-3}
\cmidrule(r){5-7}
Layer & Filter/Stride & Output Size && Layer & Filter/Stride & Output Size \\ \midrule
&&&& FC & & $6\times6\times320$ \\
Conv11 & $3\times3/1$ & $96\times96\times32$ && FConv52& $3\times3/1$ & $6\times6\times160$ \\
Conv12 & $3\times3/1$ & $96\times96\times64$ && FConv51& $3\times3/1$ & $6\times6\times256$ \\\midrule
Conv21 & $3\times3/2$ & $48\times48\times64$ && 
FConv43& $3\times3/2$ & $12\times12\times256$&&\\ 
Conv22 & $3\times3/1$ & $48\times48\times64$ &&
FConv42& $3\times3/1$ & $12\times12\times128$ \\
Conv23 & $3\times3/1$ & $48\times48\times128$ &&
FConv41& $3\times3/1$ & $12\times12\times192$ \\
\midrule
Conv31 & $3\times3/2$ & $24\times24\times128$ &&
FConv33& $3\times3/2$ & $24\times24\times192$ && \\ 
Conv32 & $3\times3/1$ & $24\times24\times96$ && FConv32& $3\times3/1$ & $24\times24\times96$ \\
Conv33 & $3\times3/1$ & $24\times24\times192$ && FConv31& $3\times3/1$ & $24\times24\times128$ \\
\midrule
Conv41 & $3\times3/2$ & $12\times12\times192$ &&
FConv23& $3\times3/2$ & $48\times48\times128$ \\ 
Conv42 & $3\times3/1$ & $12\times12\times128$ && FConv22& $3\times3/1$ & $48\times48\times64$ \\
Conv43 & $3\times3/1$ & $12\times12\times256$ && FConv21& $3\times3/1$ & $48\times48\times64$ \\
\midrule
Conv51 & $3\times3/2$ & $6\times6\times256$ && 
FConv13& $3\times3/2$ & $96\times96\times64$ \\ 
Conv52 & $3\times3/1$ & $6\times6\times160$ && FConv12& $3\times3/1$ & $96\times96\times32$ \\
Conv53 & $3\times3/1$ & $6\times6\times(N^f${\color{blue}$+1$}) && FConv11& $3\times3/1$ & $96\times96\times3$ \\
\midrule
AvgPool& $6\times6/1$ & $1\times1\times(N^f${\color{blue}$+1$}) && \\ \midrule
FC ($D$ only) && $N^d+N^p+1$ \\ \bottomrule
\end{tabular}}\figvspace 
\end{center}
\end{table}

The network structure of single-image DR-GAN is shown in Tab.~\ref{tab:network}.
We adopt CASIA-Net~\cite{yi2014learning} with batch normalization~(BN) for $G_{enc}$ and $D$. Besides, since the stability of the GAN game suffers if sparse gradient layers~(MaxPool, ReLU) are used, we replace them with  strided convolution and exponential linear unit~(ELU) respectively.
$D$ is trained to optimize Eqn.~\ref{eqn:objD} by adding a fully connected layer with the softmax loss for real vs.~fake, identity, and pose classifications respectively. 
$G$ includes $G_{enc}$ and $G_{dec}$ that are bridged by the to-be-learned identity representation $f(\mathbf{x})\in \mathbb{R}^{N^f}$, which is the AvgPool output in our $G_{enc}$.
$f(\mathbf{x})$ is concatenated with a pose code $\mathbf{c}$ and a random noise $\mathbf{z}$. 
A series of fractionally-strided convolutions (FConv)~\cite{radford2015unsupervised}
transforms the $(N^f+N^p+N^z)$-dim concatenated vector into a synthetic image $\hat{\mathbf{x}} = G(\mathbf{x}, \mathbf{c}, \mathbf{z})$, which is the same size as $\mathbf{x}$. 
$G$ is trained to maximize Eqn.~\ref{eqn:objG} when a synthetic face $\hat{\mathbf{x}}$ is fed to $D$ and the gradient is back-propagated to update $G$. 

Previous work in face rotation use $L2$ loss~\cite{zhu2014multi,yim2015rotating} to enforce the synthetic face to be similar to the ground truth face at the target pose.
This line of work requires the training data to include face image pairs of the same identity at different poses, which is achievable for controlled datasets such as Multi-PIE, but hard to fulfill for in-the-wild datasets. 
On contrary, DR-GAN does not require image pairs since there is no direct supervision on the synthetic images. 
This enables us to utilize extensive real-world unstructured datasets for model training. 
To initialize the training, given a training image, we randomly sample the pose code with equal probability for each pose view. 
Such a random sampling is conducted at {\it each} epoch during the training, for the purpose of assigning  {\it multiple} pose codes to one training image.
For the noise vector, we also randomly sample each dimension independently from the uniform distribution in the range of [$-1,1$].

%% file: PAMI_2017_DR_GAN_alg_3.3.tex
\SubSection{Multi-Image DR-GAN} 

Our single-image DR-GAN extracts an identity representation and performs face rotation by processing one single image. 
Yet, we often have multiple images per subject in training and sometimes in testing.
To leverage them, we propose multi-image DR-GAN that can benefit both the training and testing stages. 
For training, it can learn a better identity representation from multiple images that are complementary to each other. 
For testing, it can enable template-to-template matching, which addresses a crucial need in real-world surveillance applications. 

\begin{figure}[t!]
\centering
\includegraphics[trim=0 0 0 10,clip, width=0.9\linewidth]{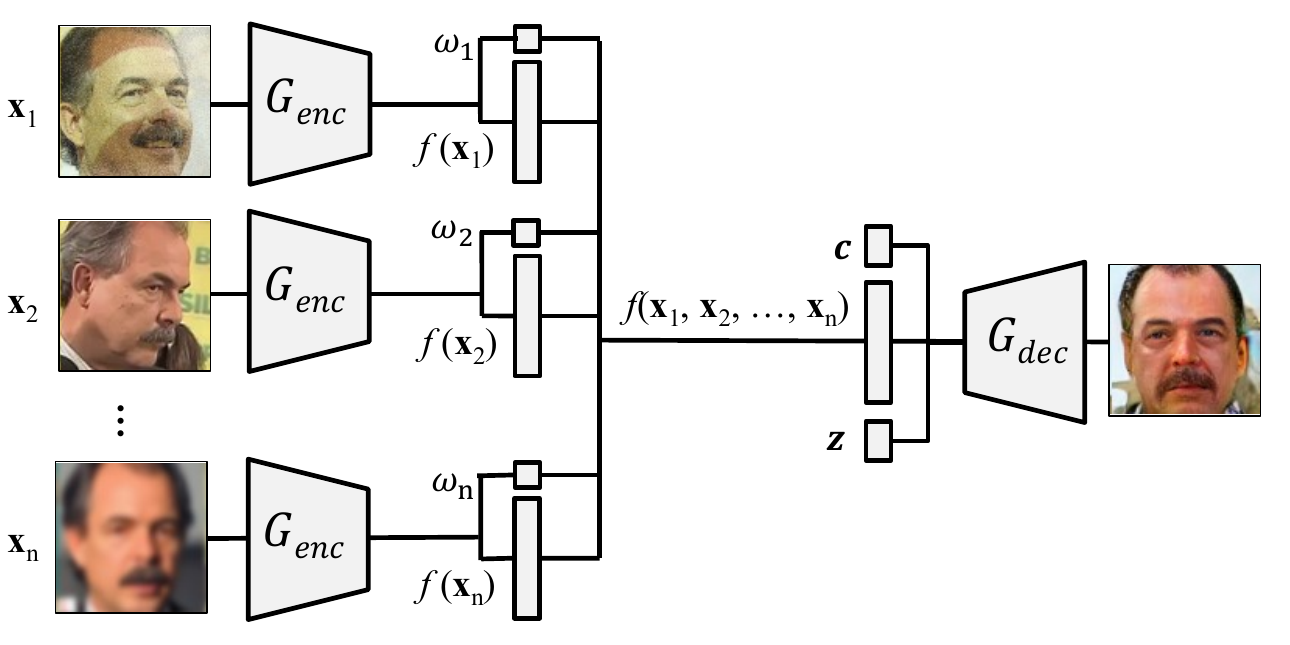}
\figvspace
\caption{\small Generator in mlti-image DR-GAN. From an image set of a subject, we can fuse the features to a single representation via dynamically learnt coefficients and synthesize images in any pose.}
\label{fig:multiple_images_model}
\figvspace
\end{figure}

The multi-image DR-GAN has the same $D$ as single-image DR-GAN, but a different $G$ as shown in Fig.~\ref{fig:multiple_images_model}. 
Given $n$ images $\{\mathbf{x}_i\}_{i=1}^n$ of the same identity $y^d$ at various poses as input, besides extracting the feature representation $f(\mathbf{x}_i)$, $G_{enc}$ also estimates a confident coefficient $\omega_i$ for each image, which predicts the quality of the learnt representation. 
The fused representation of $n$ images is the weighted average of all representations, 
\eqnvspace
\begin{equation}
f(\mathbf{x}_1, ..., \mathbf{x}_n) = \frac{\sum_{i=1}^n \omega_i f(\mathbf{x}_i)}{\sum_{i=1}^n \omega_i}. 
\label{eqn:fuse}
\end{equation}

This fused representation is then concatenated with $\mathbf{c}$ and $\mathbf{z}$ and fed to $G_{dec}$ to generate a new image, which is expected to have the same identity as all input images and a target pose $y^t$ specified by the pose code. 
Thus, each sub-objective for learning $G$ has $(n+1)$ terms:
\begin{align}
\mathcal{L}_{gan}^{G} =  \sum_{i=1}^n \Big[ {}&\mathbb{E}[\log(D^{r}(G(\mathbf{x}_i, \mathbf{c}, \mathbf{z})))] \Big] \nonumber \\ 
+ {}&  \mathbb{E}[\log(D^{r}(G(\mathbf{x_1},...,\mathbf{x}_n, \mathbf{c}, \mathbf{z})))] .
\label{eqn:objG_multi}
\end{align}

The similar extension applied for $\mathcal{L}_{id}^{G}$ and $\mathcal{L}_{pos}^{G}$.
The coefficient $\omega_i$ in Eqn.~\ref{eqn:fuse} is learned so that an image with a higher quality contributes more to the fused representation.
The quality is an indicator of the PIFR performance of the image, rather than the low-level image quality.
Face quality prediction is a classic topic where many prior work attempt to estimate the former from the latter~\cite{ozay2009improving,wong2011patch}.
Our coefficient learning is essentially the quality prediction, from novel perspectives in contrast to prior work. 
That is, without explicit supervision, it is driven by $D$ through the decoded image $G_{dec}(f(\mathbf{x}_1, ..., \mathbf{x}_n),\mathbf{c},\mathbf{z})$, and learned in the context of, as a byproduct of, representation learning.
Note that, jointly training multiple images per subject results in {\it one}, but not multiple, generator, i.e., all $G_{enc}$ in Fig.~\ref{fig:multiple_images_model} share the same parameters. 
This makes it flexible to take an {\it arbitrary number} of images during testing for representation learning and face rotation. 



For the network structure, multi-image DR-GAN only makes minor modification from the single-image counterpart.
Specifically, at the end of $G_{enc}$, we add one more convolutional filter to the layer before AvgPool to estimate the coefficient $\omega$. 
We apply $Sigmoid$ activation to constrain $\omega$ in the range of [$0,1$]. 
During training, despite unnecessary,  we keep the number of input images per subject $n$ the same for the sake of convenience in image sampling and network training.
To mimic the variation in the number of input images, we use a simple but effective trick: applying Dropout on the coefficients $\omega$: each $\omega$ is set to $0$ with a probability of $0.5$. 
Hence, during training, the network takes any number of inputs varying from $1$ to $n$. 

DR-GAN can be used in PIFR, image quality prediction, and face rotation. 
While the network in Fig.~\ref{fig:gans} (d) is used for training, our network for testing is much simplified. 
First, for PIFR, only $G_{enc}$ is used to extract the representation from one or multiple images. 
Second, for quality prediction, only $G_{enc}$ is used to compute $\omega$ from one image.
Thirdly, both $G_{enc}$ and $G_{dec}$ are used for face rotation by specifying a target pose and a noise vector.

%% file: PAMI_2017_DR_GAN_alg_3.4.tex
\subsection{Improving $G_{enc}$ via Model Switch}
\label{sec:switch}
The ultimate goal of DR-GAN is to learn a disentangled representation for PIFR. 
Our $G_{enc}$ aims for identity representation learning. 
While our $D^d$ aims for identity classification, it also learns an identity representation that could be used for face recognition during testing, the same as most previous work~\cite{yi2014learning, yin2017multi}.
The fact that both $G_{enc}$ and $D^{d}$ can be used for face recognition motivates us to explore two questions. 
First, whether $G_{enc}$ can outperform $D^d$ for PIFR.  
Second,  whether a better $D^d$ will lead to a better $G_{enc}$ in representation learning. 

\begin{figure}[t!]
\centering
\includegraphics[trim=60 228 60 240,clip, width=0.65\linewidth]{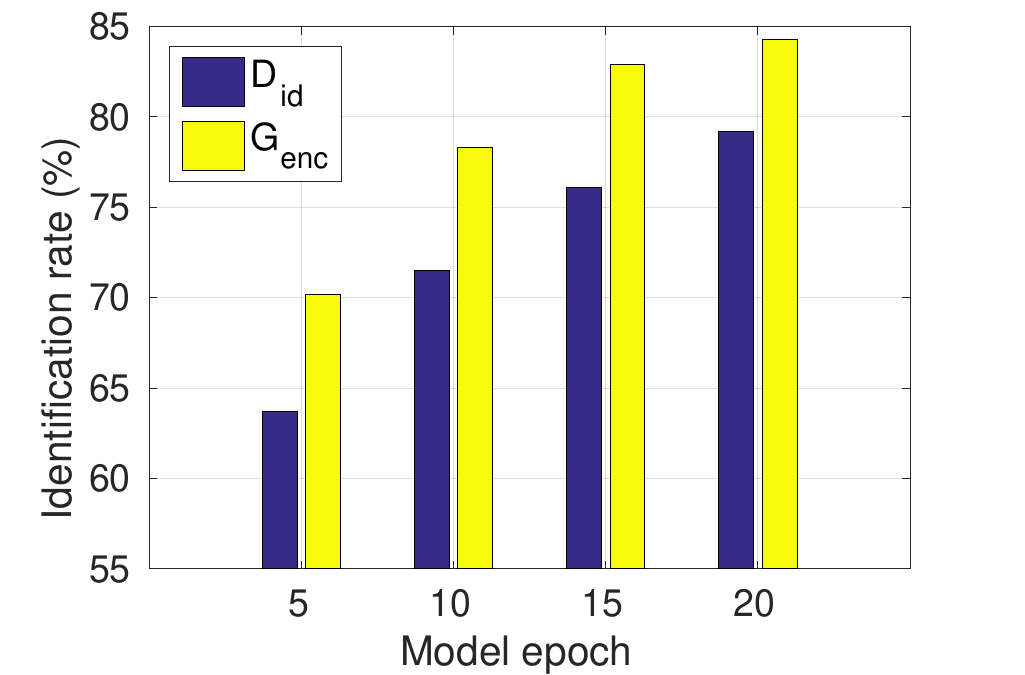}
\caption{\small Recognition performance of $G_{enc}$ and $D^{d}$ when training DR-GAN with different $D^{d}$ on Multi-PIE dataset.}
\label{fig:bounding}
\figvspace 
\end{figure}

To answer the above questions, we conduct a bounding experiment to compare the face recognition performance of $G_{enc}$ and $D^d$. 
Specifically, using the Multi-PIE training set, we train a single-task CNN-based face recognition model for $20$ epochs. 
We save the models at $5$th, $10$th, $15$th, and $20$th epochs, termed as $D^d_{5}$, $D^d_{10}$, $D^d_{15}$, $D^d_{20}$ respectively.
These four models can be used as $D^d$ and to train four single-image DR-GAN models. 
Each model is trained until converged where we only update $G$ with $D^d$ being fixed, which leads to four $G_{enc}$ termed as $G_{enc}^{5}, G_{enc}^{10}, G_{enc}^{15}, G_{enc}^{20}$ respectively.

Both $G_{enc}$ and $D^d$ are used to extract identity features for face recognition on Multi-PIE, with the results in Fig.~\ref{fig:bounding}. 
We have three observations. 
First, the performance of $D^d$ shows that $D^d_{5} <D^d_{10} < D^d_{15} < D^d_{20}$. 
This is expected since the performance increases as the model is being trained for more epochs. 
Second, the performance of $G_{enc}$ also shows a similar trend with $G_{enc}^{5} <G_{enc}^{10} <G_{enc}^{15} <G_{enc}^{20}$, which indicates that a better $D^d$ indeed leads to a better $G_{enc}$. 
Third, $G_{enc}$ consistently outperforms $D^d$, which suggests that the learnt representation in $G_{enc}$ is more discriminative than the representation in conventional CNN-based face recognition models. 

Based on the above observations, we propose an iterative scheme to switch between $G_{enc}$ and $D^d$ in order to further improve $G_{enc}$. 
As shown in Tab.~\ref{tab:network},  $G_{enc}$ and $D^d$ shares the same network structure except that $G_{enc}$ has an additional convolutional filter for the coefficient estimation. 
During training, we iteratively replace $D^d$ with the latest $G_{enc}$ by removing the additional convolutional filter after several epochs. 
Since $G_{enc}$ can always outperform $D^d$, we will expect a better $D^d$ after model switch. 
Moreover, a better $D^d$ will lead to a better $G_{enc}$, which is then used as $D^d$ for the next switch. 
This iterative switch will lead to a better representation and thus better PIFR performance.

\subsection{Improving $G_{dec}$ via Representation Interpolation}
\label{sec:improve_Gdec}
Our $G_{enc}$ learns a mapping from the image space to a representation space and $G_{dec}$ learns the mapping from the representation space to the image space. 
$G_{enc}$ is important for PIFR while $G_{dec}$ is crucial for face synthesis. 
The usage of pose code, random noise, as well as the model switch techniques are useful for learning a better disentangled representation for $G_{enc}$. 
However, even with a perfect representation from $G_{enc}$, a poor $G_{dec}$ may synthesize unsatisfactory face images. 

To learn a better $G_{dec}$, we propose to employ representation interpolation to regularize the learning process. 
Prior GANs~\cite{radford2015unsupervised} have observed that interpolation between two noise vectors can still produce a valid image. 
Similarly in our work, by assuming a convex identity space, the interpolation between two representations $f(\mathbf{x}_1)$, $f(\mathbf{x}_2)$ extracted from the face images $\mathbf{x}_1$, $\mathbf{x}_2$ of two different identities should still be a valid face but with an unknown identity. 
During training, we randomly pair images with different identities to generate an interpolated representation:
\begin{equation}
f_{\alpha}(\mathbf{x}_1, \mathbf{x}_2) = \alpha f(\mathbf{x}_1) + (1-\alpha) f(\mathbf{x}_2). 
\end{equation}

We use the average, $f_{\frac{1}{2}}$, for simplicity. 
Other fixed or random weights can be used as well. 
Similar to the objectives for $G$ and $D$ in multi-image DR-GAN, we have additional terms to regularize the averaged representation. 
$D$ aims to classify the generated image to the fake class by having the following extra term:
\begin{equation}
\mathbb{E}[\log(1 - D^{r}(G_{dec}(f_{\frac{1}{2}}(\mathbf{x_1},\mathbf{x}_2), \mathbf{c}, \mathbf{z})))]. 
\end{equation}

And $G$ aims to generate an image that can fool $D$ to classify it as the real class and the target pose, and ignore the identity part, with two additional terms in $\mathcal{L}_{gan}^{G}$ and $\mathcal{L}_{pos}^{G}$:
\begin{equation}
\mathbb{E}[\log(D^{r}(G_{dec}(f_{\frac{1}{2}}(\mathbf{x_1},\mathbf{x}_2), \mathbf{c}, \mathbf{z})))], 
\end{equation}
\begin{equation}
\mathbb{E}[\log(D^{p}_{y^t}(G_{dec}(f_{\frac{1}{2}}(\mathbf{x_1},\mathbf{x}_2), \mathbf{c}, \mathbf{z})))]. 
\end{equation}
 

With the proposed techniques to improve both $G_{enc}$ and $G_{dec}$, we expect to improve the generalization ability of DR-GAN for both representation learning and image synthesis. 
As will be shown in the experiments, the proposed techniques are effective in improving the performance of DR-GAN.

\begin{figure*}[t!]
\begin{center}
\includegraphics[width=0.99\linewidth]{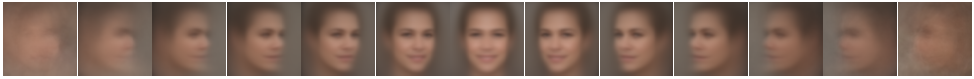}
\caption{The mean faces of $13$ pose groups in CASIA-Webface. The blurriness shows the challenges of pose estimation for large poses.}
\label{fig:mean_faces} \vspace{-2mm}
\figvspace
\end{center}
\end{figure*}

%% file: PAMI_2017_DR_GAN_alg_3.5.tex
\SubSection{Comparison to Prior GANs}
\label{sec:gans}
We compare DR-GAN with most relevant GAN variants (Fig.~\ref{fig:gans}).

\Paragraph{Conditional GAN} Conditional GAN~\cite{mirza2014conditional, kwak2016ways} extends GAN by feeding the labels to both $G$ and $D$ to generate images conditioned on  labels, either class labels, modality information, or even partial data for inpainting. 
It has been used to generate MNIST digits conditioned on the class label and to learn multi-modal models. 
In conditional GAN, $D$ is trained to classify a real image with mismatched conditions to a fake class. 
In DR-GAN, $D$ classifies a real image to the corresponding class based on the labels. 


\Paragraph{Auxiliary Classifier GAN} Odena~\etal~\cite{odena2017conditional} extends conditional GAN to add an additional classifier to $D$ to classify real images into $N^c$ classes. 
DR-GAN shares a similar loss for $D$ but with a distinguish purpose. The auxiliary classifier in Odena\etal~\cite{odena2017conditional} is used to help improving the stability and quality of GAN training. Meanwhile, we employ two additional classifiers to guide the representation learning in the encoder-decoder structure $G$.

\Paragraph{Adversarial Autoencoder (AAE)}
In AAE~\cite{makhzani2015adversarial}, $G$ is the encoder of an autoencoder. 
AAE has two objectives in order to turn an autoencoder into a generative model: the autoencoder reconstructs the input image, and the latent vector generated by the encoder matches an arbitrary prior distribution by training $D$. 
DR-GAN differs to AAE in two aspects. 
First, the autoencoder in~\cite{makhzani2015adversarial} is trained to learn a latent representation similar to an imposed prior distribution, while our encoder-decoder learns discriminative identity representations. 
Second, $D$ in AAE is trained to distinguish real/fake distributions while our $D$ is trained to classify real/fake images, the identity and pose of the images. 

%% file: PAMI_2017_DR_GAN_exp.tex
\Section{Experiments}
DR-GAN can be used for face recognition by using the learnt representation from $G_{enc}$, and face rotation by specifying different pose codes and noise vectors with $G$. 
We evaluate DR-GAN quantitatively for PIFR and qualitatively for face rotation. 
We further conduct experiments to analyze the training strategy, disentangle representation, and image coefficients.  
Our experiments are conducted for both controlled and in-the-wild databases. 

\SubSection{Experimental Settings}
\Paragraph{Databases}
\label{sec:dataset}
Multi-PIE~\cite{gross2010multi} is the largest database for evaluating face recognition under pose, illumination, and expression variations in controlled setting. 
For fair comparison, we follow the setting in~\cite{zhu2014multi}: using $337$ subjects with neutral expression, $9$ poses within $\pm60^\circ$, and $20$ illuminations. 
The first $200$ subjects are used for training and the rest $137$ subjects for testing. 
In the testing set, one image per subject with frontal view and neutral illumination forms the gallery set and the others are the probe set. 
For Multi-PIE experiments, we add an additional illumination code similar to the pose code to disentangle the illumination variation. 
Therefore, we have $N^d=200$, $N^p=9$, $N^{il}=20$.
Further, to demonstrate our ability in synthesizing large-pose faces, we train a second model with training faces up to $90^\circ$ (i.e., $N^p=13$).

For the in-the-wild setting, we train on CASIA-WebFace~\cite{yi2014learning} and AFLW~\cite{koestinger2011annotated}, and test on CFP~\cite{sengupta2016frontal} and IJB-A~\cite{klare2015pushing}.  
CASIA-WebFace includes $494,414$ near-frontal faces of $10,575$ subjects. 
We add the AFLW ($25,993$ images) to the training set to supply more pose variation. Since there is no identity information in this dataset, those images only used to compute GAN, pose related losses.
CFP consists of $500$ subjects each with $10$ frontal and $4$ profile images. 
The evaluation protocol includes frontal-frontal (FF) and frontal-profile (FP) face verification, each having $10$ folders with $350$ same-person pairs and $350$ different-person pairs.
As another large-pose database, IJB-A has $5,396$ images and $20,412$ video frames of $500$ subjects. 
It defines template-to-template face recognition where each template has one or multiple images. 
We remove $27$ overlap subjects between CASIA-Webface and IJB-A  from the training.
We have $N^d=10,548$, $N^p=13$. 
We set $N^{f}=320$, $N^{z}=50$ for both settings.

\Paragraph{Implementation Details}
Following~\cite{yi2014learning}, we align all face images to a canonical view of size $110\times110$.
We randomly sample $96\times 96$ regions from the aligned $110\times 110$ face images for data augmentation. 
Image intensities are linearly scaled to the range of $[-1, 1]$. 
To provide pose labels $y^p$ for CASIA-WebFace, we apply $3$D face alignment~\cite{jourabloo2017pose,jourabloo2017pose-ijcv} to classify each face to one of $13$ poses. 
The mean face image of each pose group is shown in Fig.~\ref{fig:mean_faces}.
The mean faces of profile faces are less sharp than those of the near-frontal pose groups, which indicates the pose estimation error caused by the face alignment algorithm. 

Our implementation is extensively modified from a publicly available implementation of DC-GAN. 
We follow the optimization strategy in~\cite{radford2015unsupervised}.
The batch size is set to be $64$. 
All weights are initialized from a zero-centered normal distribution with a standard deviation of $0.02$. 
Adam optimizer~\cite{kingma2014adam} is used with a learning rate of $0.0002$ and momentum $0.5$.

\Paragraph{Evaluation}
The proposed DR-GAN aims for both face representation learning and face image synthesis. 
The cosine distance between two representations is used for face recognition. 
We also evaluate the performance of face recognition w.r.t.~different numbers of images in both training and testing. 
For image synthesis, we show qualitative results by comparing different losses and interpolation of the learnt representations. 
We also evaluate the various effects of different components in our method.

\begin{table}[t!]
\caption{\small{DR-GAN and its partial variants performance comparison.}}
\vspace{-3mm}
\small
\begin{center}
\begin{tabular}{@{\hskip 1.5mm}l@{\hskip 1.5mm}c@{\hskip 1.5mm}c@{\hskip 1.5mm}c@{\hskip 1.5mm}c@{\hskip .5mm}}
\toprule
& \multicolumn{2}{c}{Verification} & \multicolumn{2}{c}{Identification} \\ \cmidrule(r){2-3} \cmidrule(r){4-5}
Method & @FAR=$.01$ & @FAR=$.001$ & @Rank-$1$ & @Rank-$5$ \\ \midrule
DR-GAN $- D^r$ & $80.0\pm2.2$ & $55.5\pm3.5$ & $88.7\pm0.8$ & $95.0\pm0.8$ \\
DR-GAN $- D^p$ & $78.0\pm2.0$ & $53.9\pm6.8$ & $87.5\pm0.8$ & $94.5\pm0.7$ \\
DR-GAN         & $81.2\pm2.7$ & $56.2\pm9.1$ & $89.0\pm1.4$ & $95.1\pm0.9$ \\
 \bottomrule
\end{tabular}
\end{center}
\eqnvspace
\label{tab:ijb-a_D_components}
\vspace{-2mm}
\end{table}

\begin{figure}[t!]
\begin{center}
\includegraphics[trim=0 0 0 0,clip,width=\linewidth]{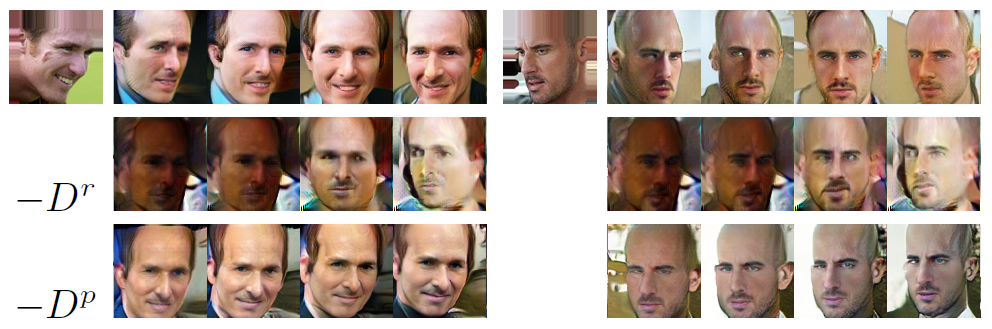}
\end{center}
\vspace{-3mm}
\caption{Generated faces of DR-GAN and its partial variants.}
\label{fig:ijb-a_D_components_vis}
\figvspace
\end{figure}

\SubSection{Ablation study}
\label{sec:ablation}
\Paragraph{Discriminator Components}
\label{sec:ablation_D_components}
Our discriminator is designed as a multi-task CNN with three components, namely $D^g, D^d, D^p$, for real/fake, identity and pose classification respectively. 
While $D^d$ plays a critical role to guide the generator to preserve the input identity, we would like to study the role of the remaining components.
Table~\ref{tab:ijb-a_D_components} presents the recognition performance of single-image DR-GAN partial variants with each of $D$ components removed. 
While the variant without adversarial loss has a slightly performance drop, the model without pose classification task has more severe drop. 
This shows the important of generating face images in different poses. 
Also, the role of each component is shown in generated faces (Fig.~\ref{fig:ijb-a_D_components_vis}). 
When removing $D^r$, generated images has lower quality although they can be realized as faces and in correct poses. 
When removing $D^p$, the pose of generated images can't be controlled by the pose code and usually affected by the input face's pose. 
This can be caused by pose information residing in the feature representation. 
This also explains the severe drop in the model's recognition performance.

\Paragraph{Disentangled Representation}
\label{sec:ablation_disentangled_representation}
In DR-GAN, we claim that the learnt representation is disentangled from pose variations via the pose code. 
To validate this, following the energy-based weight visualization method proposed in~\cite{yin2017multi}, we perform feature visualization on the FC layer, denoted as $\mathbf{h}\in \mathbb{R}^{ 6\times6\times320}$, in $G_{dec}$.
Our goal is to select two out of the $320$ filters that have highest responses for identity and pose respectively. 
The assumption is that if the learnt representation is pose-invariant, there should be separate neurons to encode the identity features and pose features. 

Recall that we concatenate $f(\mathbf{x})\in \mathbb{R}^{320}$, $\mathbf{c}\in \mathbb{R}^{13}$ and $\mathbf{z}\in\mathbb{R}^{50}$ into one feature vector, which multiplies with a weight matrix $\mathbf{W}_{fc} \in \mathbb{R}^{(320+13+50)\times(6\times6\times320)}$ and generates the output $\mathbf{h}$ with $\mathbf{h}^{i}\in\mathbb{R}^{6\times6}$ being the feature output of one filter in FC. 
Let $\mathbf{W}_{fc} = [\mathbf{W}_{fx}; \mathbf{W}_{c}; \mathbf{W}_{z}]$ denote the weight matrix with three sub-matrices, which would multiply with $f(\mathbf{x}), \mathbf{c}, \mathbf{z}$ respectively. 
Taking the identity matrix as an example, we have $\mathbf{W}_{fx}=[\mathbf{W}_{fx}^1, \mathbf{W}_{fx}^2, ..., \mathbf{W}_{fx}^{320}]$ where $\mathbf{W}_{fx}^i\in\mathbb{R}^{320\times36}$. 
We compute an energy vector $\mathbf{s}_d\in\mathbb{R}^{320}$ with each element as: $\mathbf{s}_{d}^i = || \mathbf{W}_{fx}^i ||_F$.
We then find the filter with the highest energy in $\mathbf{s}_d$ as $k_d =\operatorname*{arg\,max}_i \mathbf{s}_d^i$.
Similarly, by partitioning $\mathbf{W}_c$, we find another filter, denoted as $k_p$, with the highest energy for pose. 

\begin{figure}[t!]
  \begin{center}
  \small
   \includegraphics[width=0.98\linewidth]{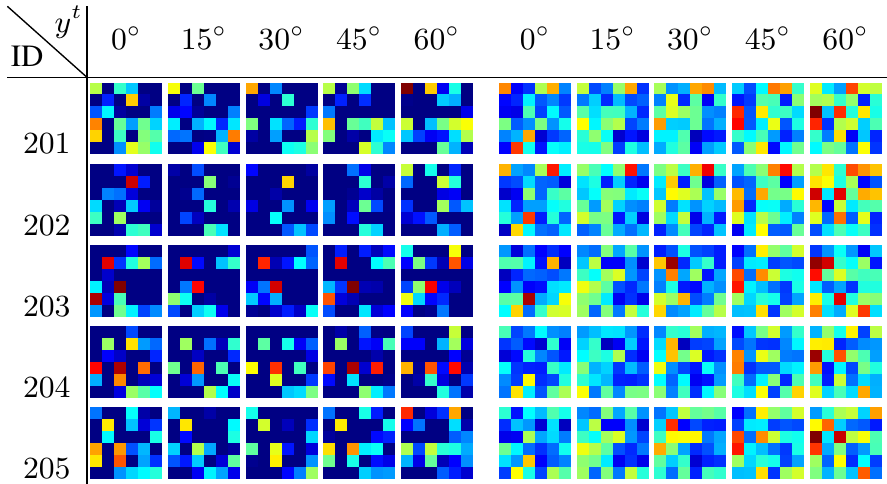}
\vspace{-2mm}
\caption{\small Responses of two filters: filter with the highest responses to identity (left), and pose (right). Responses of each row are of the same subject, and each column are of the same pose. Note the within-row similarity on the left and within-column similarity on the right.}
\label{fig:feature_visual}\figvspace \vspace{-1mm}
\end{center}
\end{figure}

Given the representation $f(\mathbf{x})$ of one subject, along with a pose code $\mathbf{c}$ and noise $\mathbf{z}$, we can compute the responses of two filters via $\mathbf{h}^{k_d} = (f(\mathbf{x}); \mathbf{c}; \mathbf{z})^\T\mathbf{W}_{fc}^{k_d}$ and $\mathbf{h}^{k_p} = (f(\mathbf{x}); \mathbf{c}; \mathbf{z})^\T\mathbf{W}_{fc}^{k_p}$.
By varying the subjects and pose codes, we generate two arrays of responses in Fig.~\ref{fig:feature_visual}, for identity ($\mathbf{h}^{k_d}$) and pose ($\mathbf{h}^{k_p}$) respectively.
For both arrays, each row represents the responses of the same subject and each column represents the same pose. %
The responses for identity encode the identity features, where each row shows similar patterns and each column does not share similarity. 
On contrary, for pose responses, each column share similar patterns while each row is not related. 
This visualization supports our claim that the learnt representation is pose-invariant.

\Paragraph{Single vs. Multiple Image DR-GAN}
\label{sec:single_vs_multiple_training}
We evaluate the effect of the number of training images ($n$) per subject on the face recognition performance on CFP. 
Specifically, with the {\it same} training set, we train three models with $n=1,4,6$, where $n=1$ denotes single-image DR-GAN and $n>1$ denotes multi-image DR-GAN.
The face verification performance on CFP using $f({\bf{x}})$ of each model are shown in Tab.~\ref{tab:CFP_results_1}.
We observe the advantage of multi-image DR-GAN over the single-image counterpart despite they use the {\it same amount} of training data, which attributes to more constraints in learning $G_{enc}$ that leads to a better representation. 
However, we do not keep increasing $n$ due to the limited computation capacity. 
In the rest of the paper, we use multi-image DR-GAN with $n=6$ unless specified. 

\begin{table}[t!]
\small
\caption{\small{Comparison of single vs.~multi-image DR-GAN on CFP.}} 
\vspace{-3mm}
\label{tab:CFP_results_1}
\begin{center}
\begin{tabular}{ lcc}
\toprule 
Method & Frontal-Frontal & Frontal-Profile\\ \midrule
DR-GAN: n=$1$ & $97.13\pm0.68$ & $90.82\pm0.28$ \\ 
DR-GAN: n=$4$ & $97.86\pm0.75$ & $92.93\pm1.39$ \\ 
DR-GAN: n=$6$ & $97.84\pm0.79$ & $93.41\pm1.17$ \\ 
\bottomrule
\end{tabular}
\end{center}
\figvspace
\end{table}

\begin{table}[t]
\begin{center}
\caption{\small Performance of  $G_{enc}$ on Multi-PIE when keep switching to $D_{d}$. At Epoch $0$, $G_{enc}$ is trained with only the softmax loss. }
\label{tab:boosting}
\small
\resizebox{0.99\linewidth}{!}{
\begin{tabular}{ lcccccccccc }
\toprule
 Epoch No.                & $0$ & $20$ & $40$ & $60$ & $80$ & $100$ \\ \midrule
 Identification rate (\%) &  $79.7$ & $84.8$ & $87.1$ & $88.7$ & $89.8$ & $90.4$ \\
\bottomrule
\end{tabular}}
\end{center}\figvspace\vspace{-2mm}
\end{table}

\Paragraph{Model Switch}
In Sec.~\ref{sec:switch}, we propose to improve $G_{enc}$ via model switch, i.e., replacing $D^d$ with $G_{enc}$ during training. 
Table~\ref{tab:boosting} shows the performance of $G_{enc}$ for face recognition on Multi-PIE. 
At the beginning, $G_{enc}$ is initilized with a model trained with the softmax loss for identity classification. 
We use $G_{enc}$ to replace $D^d$ and retrain $G$ with random initialization.
When $G$ converges, based on the accuracy on a $5$-held-out-subjects validation set, we replace $D^d$ with $G_{enc}$ and repeat above steps. 
Table~\ref{tab:boosting} reports face recognition performance of $G_{enc}$ on Multi-PIE test set at each switch. 
Clearly, the performance keeps improving as training goes on.
This study implies that DR-GAN may leverage the future development of face recognition, by using a $3$rd party recognizer as $D^d$ and further improve upon it.

\SubSection{Confident Coefficients}

\begin{figure*}[t!]
\begin{center}
\includegraphics[trim=0 0 70 170, clip, width=0.98\linewidth]{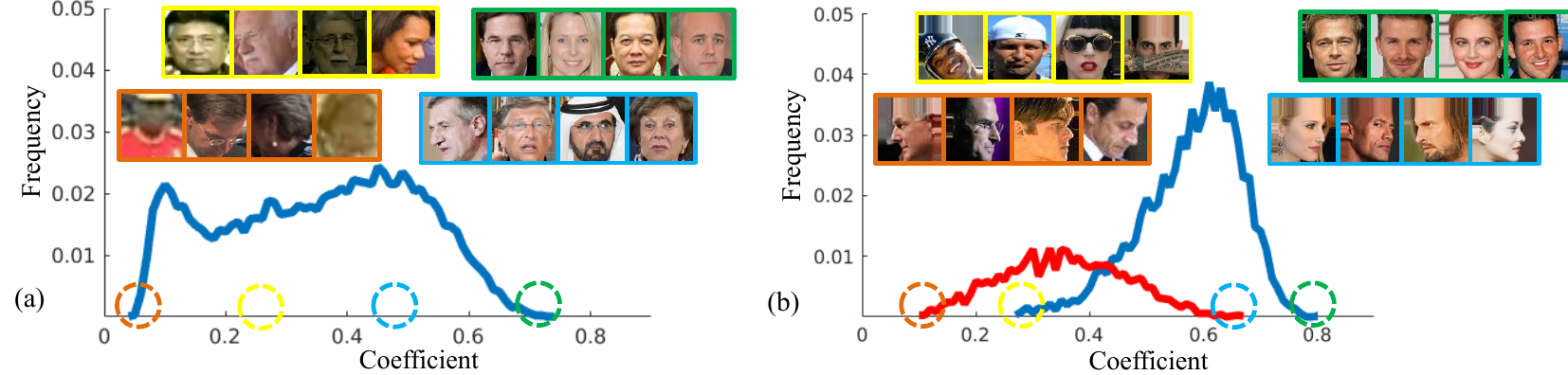}
\end{center}
\vspace{-1mm}
   \caption{\small Coefficient distributions on IJB-A (a) and CFP (b). For IJB-A, we visualize images at four regions of the distribution. For CFP, we plot the distributions for frontal faces (blue) and profile faces (red) separately and show images at the heads and tails of each distribution. }
\label{fig:coeff}
\end{figure*}

In multi-image DR-GAN, we learn a confident coefficient for each input image by assuming that the learnt coefficient is indicative of the image quality, i.e., how good it can be used for face recognition. 
Therefore, a low-quality image should have a relatively poor representation and small coefficients so that it would contribute less to the fused representation.  
To validate this assumption, we compute the confident coefficients for all images in IJB-A and CFP databases and plot the distribution as shown in Fig.~\ref{fig:coeff}.

For IJB-A, we show four example images with low, medium-low, medium-high, and high coefficients. 
It is obvious that the learnt coefficients are correlated to the image quality. 
Images with relatively low coefficients are usually blurring, with large poses or failure cropping. 
While images with relatively high coefficients are of very high quality with frontal faces and less occlusion. 
Since CFP consists of $5,000$ frontal faces and $2,000$ profile faces, we plot their distributions separately.
Despite some overlap in the middle region, the profile faces clearly have relatively low coefficients compared to the frontal faces. 
Within each distribution, the coefficient are related to other variations expect yaw angles. 
The low-quality images for each pose group are with occlusion and/or challenging lighting conditions, while the high-quality ones are with less occlusion and under normal lighting. 

\begin{figure}[t!]
\begin{center}
\includegraphics[trim=115 230 145 240, clip, width=0.5\linewidth]{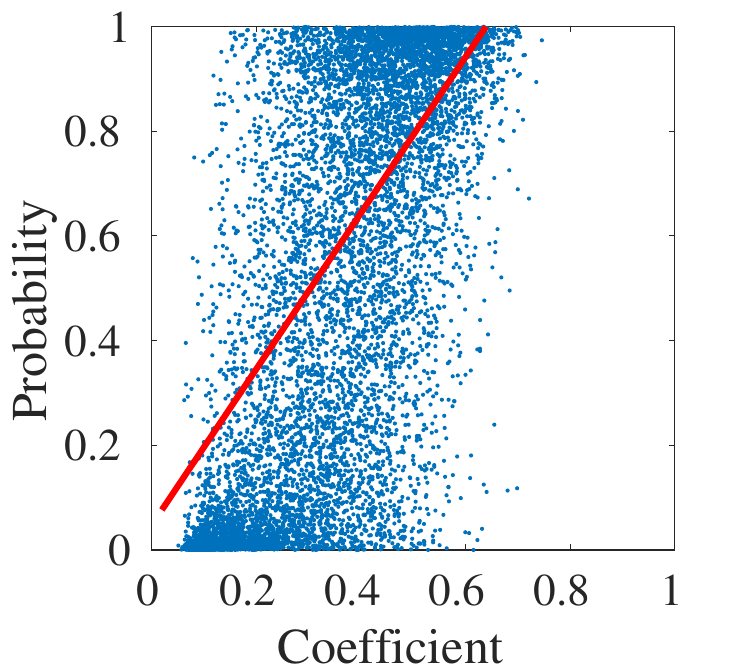}
\end{center}
\vspace{-1mm}
\caption{The correlation between the estimated coefficients and the classification probabilities.}
\label{fig:coeff_corr}
\end{figure}

To quantitatively evaluate the correlation between the coefficients and face recognition performance, we conduct an identity classification experiment on IJB-A.  
Specifically, we randomly select all frames of one video for each subject and select half of images for training and remaining for testing.
The training and testing sets share the same identities.
Therefore, in the testing stage, we can use the output of the softmax layer as the probability of each testing image belonging to the right identity class. 
This probability is an indicator of how well the input image can be recognized as the true identity.  
Given the estimated coefficients, we plot these two values for the testing set, as shown in Fig.~\ref{fig:coeff_corr}. 
These two values are highly correlated to each other with a correlation of $0.69$, which again supports our assumption that the learnt coefficients are indicative of the image quality. 

\Paragraph{Image selection with $\omega$}
One common application of image quality is to prevent low-quality images from contributing to face recognition. 
To validate whether our coefficients have such usability, we design the following experiment.
For each template in IJB-A, we keep images whose coefficients $\omega$ are larger than a predefined threshold $\omega_t$, or if all $\omega$ are smaller we keep one image with the highest $\omega$.
Tab.~\ref{tab:ijb-a_threshold} reports the performance on IJB-A, with different $\omega_t$. 
With $\omega_t$ being $0$, all test images are kept and the result is the same as Tab.~\ref{tab:ijb-a}. 
These results show that keeping all or majority of the samples are better than removing them.
This is encouraging as it reflects the effectiveness of DR-GAN in automatically diminishing the impact of low-quality images, without removing them by thresholding.
\begin{table}[t!]
\caption{\small{Performance of IJB-A when removing images by threshold $\omega_t$. ``Selected" shows the percentage of retained images.}}
\vspace{-3mm}
\small
\begin{center}
\begin{tabular}{@{\hskip .5mm}l@{\hskip 1.5mm}c@{\hskip 1.5mm}c@{\hskip 1.5mm}c@{\hskip 1.5mm}c@{\hskip 1.5mm}c@{\hskip .5mm}}
\toprule
\multirow{2}{*}{$\omega_t$}& Selected & \multicolumn{2}{c}{Verification} & \multicolumn{2}{c}{Identification} \\ \cmidrule(r){3-4} \cmidrule(r){5-6}
 & ($\%$) & @FAR=$.01$ & @FAR=$.001$ & @Rank-$1$ & @Rank-$5$ \\ \midrule 
$0   $ &$100.0$ & ${\mathbf{84.3}} \pm 1.4$ & $72.6 \pm 4.4$ & $91.0\pm 1.5$ & $95.6\pm1.1$ \\
$0.1 $ &$94.9$ & $84.2 \pm 1.7$ & $72.7 \pm 2.9$ & ${\mathbf{91.3}}\pm 1.3$ & ${\mathbf{95.7}}\pm1.0$ \\
$0.25$ &$71.9$ & $83.6 \pm 1.2$ & ${\mathbf{73.3}} \pm 3.0$ & $90.7\pm 1.2$ & $95.2\pm1.0$ \\
$0.5 $ &$24.6$ & $80.9 \pm 1.9$ & $71.3 \pm 4.7$ & $86.5\pm 1.9$ & $93.1\pm1.6$ \\
$1.0 $ &$5.7$ & $77.8 \pm 2.2$ & $64.0 \pm 6.2$ & $83.4\pm 2.3$ & $91.6\pm 1.2$ \\ \bottomrule
\end{tabular}
\end{center}
\label{tab:ijb-a_threshold}
\vspace{-2mm}
\end{table}

\Paragraph{Feature fusion with $\omega$}
We also would like to show our proposed feature fusion using coefficient $\omega$ is effective for the template to template matching purpose. 
We compare it with multiple fusion methods in both feature level and score level. 
Table~\ref{tab:fusion_comparison} shows comparisons of different fusion methods on our multi-image DR-GAN features. 
To compare two template with size $n_1, n_2$, for score-level, min, max, mean are respectively taking minimum, maximum and average of all $n_1n_2$ possible pairwise distances. 
Mean-min is the average of $n_1 + n_2$ minimum distances from each feature from one template to the other. 
All of these methods have the time complexity of $\mathcal{O}(n_1n_2)$. 
Softmax, proposed in~\cite{abdalmageed2016face}, aggregates multiple weighted averages of the pair-wise scores, where each weight is the function of the score using an exponential function in different scales. 
It has the time complexity of $\mathcal{O}(mn_1n_2)$, where $m$ is the number of weight scale.  
Here, following~\cite{masi2016pose}, we use a total of $m=21$ scales from $0$ to $20$.
For feature-level fusion, max, mean are respectively max-pooling and average-pooling along each feature dimension. 
All feature-level fusion methods, including our $\omega$-fusion, have the time complexity of $\mathcal{O}(n_1+n_2)$. 
From Tab.~\ref{tab:fusion_comparison}, our fusion using estimated $\omega$ achieves the best performance among all methods.

\begin{table}[t!]
\caption{\small{Fusion schemes comparisons on IJB-A dataset. }}
\vspace{-3mm}
\small
\begin{center}
\begin{tabular}{@{\hskip .0mm}l@{\hskip 1.5mm}l@{\hskip 1.5mm}c@{\hskip 1.5mm}c@{\hskip 1.5mm}c@{\hskip 1.5mm}c@{\hskip .5mm}}
\toprule
&& \multicolumn{2}{c}{Verification} & \multicolumn{2}{c}{Identification} \\ \cmidrule(r){3-4} \cmidrule(r){5-6}
& Method & @FAR=$.01$ & @FAR=$.001$ & @Rank-$1$ & @Rank-$5$ \\ \midrule
\multirow{5}{*}{\rotatebox[origin=c]{90}{Score}} 
& Min & $78.3\pm2.7$ & $46.0\pm6.9$ & $86.7\pm1.4$ & $94.0\pm0.6$  \\
& Max & $22.8\pm2.0$ & $12.3\pm2.3$ & $30.6\pm2.8$ & $52.8.0\pm2.7$\\
& Mean & $72.8\pm2.9$ & $49.2\pm5.3$ & $85.7\pm1.3$ & $93.1\pm0.6$\\
& Mean-min & $82.4\pm2.2$ & $58.5\pm6.3$ & $90.2\pm1.0$ & $\mathbf{95.6}\pm0.5$ \\
& Softmax & ${\mathbf{84.3}}\pm1.6$ & $69.2\pm6.8$ & $ 90.1\pm1.0$ & $95.5\pm0.8$ \\ \midrule
\multirow{3}{*}{\rotatebox[origin=c]{90}{Feature}} 
& Max & $19.0\pm1.3$ & $12.1\pm1.7$ & $45.4\pm5.3$ & $62.6\pm0.9$ \\
& Mean & $83.0\pm1.5$ & $67.0\pm4.8$ & $89.6\pm1.5$ & $95.4\pm0.7$ \\
& $\omega$-fusion  & ${\mathbf{84.3}}\pm1.4$ & ${\mathbf{72.6}}\pm4.4$ & ${\mathbf{91.0}}\pm1.5$ & ${\mathbf{95.6}}\pm1.1$ \\ 
 \bottomrule
\end{tabular}
\end{center}
\eqnvspace
\label{tab:fusion_comparison}
\vspace{-2mm}
\end{table}

\SubSection{Representation Learning}

\Paragraph{Loss Function Comparison}
Our $G_{dec}$ and $D$ can be viewed as a loss function for $f(\mathbf{x})$.
Typical loss functions used in deep learning-based face recognition can be divided into two categories: probability- and energy-based losses. 
Probability-based losses (i.e., softmax and its variants) usually compute a distribution of probability to all identities. 
Meanwhile, energy-based losses (contrastive, triplet, etc.) associate an energy to each configuration. 
Here, we compare DR-GAN to multiple common loss functions of face recognition. 
To have a fair comparison on IJB-A, for all functions, we use our $G_{enc}$ network architecture and ``mean min" fusion.
DR-GAN by itself can surpass all prior loss functions (Tab.~\ref{tab:loss_comparison}). 
Also, any advanced loss function can also be beneficial to DR-GAN: energy-based losses (center, triplet, etc.) can be employed directly on our representation $f(\mathbf{x})$ or probability-based losses (angular, additive-margin softmax, etc.) can be used to replace the $D_{d}$'s softmax. 
Empirically, using additive-margin softmax~\cite{wang2018additive} as a softmax replacement on $D_{d}$ can further improve DR-GAN performance, we name this variant as \DrGanAM. 

\begin{table}[t!]
\caption{\small{Loss function comparisons. All use ``mean min" fusion.}}
\vspace{-3mm}
\small
\begin{center}
\begin{tabular}{@{\hskip 1mm}l@{\hskip 1mm}c@{\hskip 1mm}c@{\hskip 1mm}c@{\hskip 1mm}c@{\hskip .5mm}}
\toprule
& \multicolumn{2}{c}{Verification} & \multicolumn{2}{c}{Identification} \\ \cmidrule(r){2-3} \cmidrule(r){4-5}
Method & @FAR=$.01$ & @FAR=$.001$ & @Rank-$1$ & @Rank-$5$ \\ \midrule
Softmax & $75.9\pm3.9$ & $44.1\pm9.9$ & $87.8\pm0.9$ & $94.6\pm0.6$ \\
Center~\cite{wen2016discriminative} & $74.9\pm3.1$ & $50.3\pm7.0$ & $87.2\pm1.4$ & $95.2\pm0.9$ \\
Triplet~\cite{schroff2015facenet} & $74.9\pm3.1$ & $50.3\pm7.0$ & $87.2\pm1.4$ & $95.2\pm0.9$ \\
AM-Softmax~\cite{wang2018additive} & $81.3\pm3.0$ & $52.7\pm8.9$ & $88.7\pm0.7$ & $94.3\pm0.4$ \\ \hline
DR-GAN$_{\text{single img.}}$ 
                 & $81.2\pm2.7$ & $56.2\pm9.1$ & $89.0\pm1.4$ & $95.1\pm0.9$ \\
DR-GAN         & $82.4\pm2.3$ & $58.5\pm8.0$ & $90.2\pm1.0$ & ${\bf{95.6}}\pm0.5$ \\
\DrGanAM  & ${\bf{85.7}}\pm1.6$ & ${\mathbf{70.3}}\pm5.79$ & ${\bf{91.0}}\pm1.5$ & ${\bf{95.6}}\pm1.1$ \\ 
 \bottomrule
\end{tabular}
\end{center}
\eqnvspace
\label{tab:loss_comparison}
\end{table}

\Paragraph{Results on Benchmark Databases}
\label{sec:benchmark}
We compare DR-GAN with state-of-the-art face recognizers on IJB-A, CFP and Multi-PIE.

\begin{table}[t!]
\caption{\small{Performance comparison on IJB-A dataset.}}
\vspace{-3mm}
\small
\begin{center}
\begin{tabular}{@{\hskip .5mm}l@{\hskip 1.5mm}c@{\hskip 1.5mm}c@{\hskip 1.5mm}c@{\hskip 1.5mm}c@{\hskip .5mm}}
\toprule
& \multicolumn{2}{c}{Verification} & \multicolumn{2}{c}{Identification} \\ \cmidrule(r){2-3} \cmidrule(r){4-5}
Method & @FAR=$.01$ & @FAR=$.001$ & @Rank-$1$ & @Rank-$5$ \\ \midrule
GOTS~\cite{klare2015pushing} & $40.6\pm1.4$ & $19.8\pm0.8$ & $44.3\pm2.1$ & $59.5\pm2.0$ \\
Wang \etal~\cite{wang2016face} & $72.9\pm3.5$ & $51.0\pm6.1$ & $82.2\pm2.3$ & $93.1\pm1.4$ \\ 
DCNN~\cite{chen2016unconstrained} & $78.7\pm4.3$ & \textendash & $85.2\pm1.8$ & $93.7\pm1.0$ \\
PAM$_{frontal}$~\cite{masi2016pose} & $73.3\pm1.8$ & $55.2\pm3.2$ & $77.1\pm1.6$ & $88.7\pm0.9$ \\
PAMs~\cite{masi2016pose} & $82.6\pm1.8$ & $65.2\pm3.7$ & $84.0\pm1.2$ & $92.5\pm0.8$ \\ 
p-CNN~\cite{yin2017multi} &  $77.5 \pm 2.5$ & $53.9 \pm 4.2$ &  $85.8 \pm 1.4$ & $93.8 \pm 0.9 $ \\
FF-GAN~\cite{yin2017towards} & $85.2 \pm 1.0$ & $66.3 \pm 3.3$ &  $90.2 \pm 0.6$ & $95.4 \pm 0.5 $ \\ \midrule

DR-GAN~\cite{tran2017disentangled} & $77.4\pm2.7$ & $53.9\pm4.3$ & $85.5\pm1.5$ & $94.7\pm1.1$ \\
DR-GAN$_{\text{split}}$ & $84.3\pm1.4$ & $72.6\pm4.4$ & $91.0\pm1.5$ & $95.6\pm1.1$ \\
DR-GAN$_{\text{split+inter}}$ & $85.6\pm1.5$ & $75.1\pm4.2$ & $91.3\pm1.6$ & $95.8\pm1.0$ \\
\DrGanAM & $\mathbf{87.2}\pm1.4$ & $\mathbf{78.1}\pm3.5$ & $\mathbf{92.0}\pm1.3$ & $\mathbf{96.1}\pm0.7$
 \\
 \bottomrule
\end{tabular}
\end{center}
\eqnvspace
\label{tab:ijb-a}
\vspace{-2mm}
\end{table}

Table~\ref{tab:ijb-a} shows the performance of both face identification and verification on IJB-A.
For our results, we report results of multi-image DR-GAN using the proposed $\omega$-fusion.
The first row shows the performance of our preliminary work~\cite{tran2017disentangled}.
``split'' represents the model trained with the separated $D^r$.
``+inter'' represents the additional changes made by the representation interpolation proposed in Sec.~\ref{sec:improve_Gdec}, which is shown to be effective in improving the face recognition performance. 
The final row presents the variant using additive margin softmax~\cite{wang2018additive} (also with ``split" and ``interpolation"). 
Compared to the state of the art, DR-GAN achieves superior results on both verification and identification. 
Also, our work has made substantial improvement over the preliminary version~\cite{tran2017disentangled}. 
These in-the-wild results show the power of DR-GAN for PIFR.

\begin{table}[t!]
\small
\caption{\small{Performance (Accuracy) comparison on CFP.}} 
\label{tab:CFP_results}
\vspace{-4mm}
\begin{center}
\begin{tabular}{ lcc}
\toprule 
Method & Frontal-Frontal & Frontal-Profile\\ \midrule
Sengupta et al.~\cite{sengupta2016frontal} & $96.40\pm0.69$ & $84.91\pm1.82$ \\
Sankarana et al.~\cite{sankaranarayanan2016triplet} & $96.93\pm0.61$ & $89.17\pm2.35$ \\ 
Chen et al.~\cite{chen2016fisher} & ${\bf{98.67}}\pm0.36$ & $91.97\pm1.70$ \\
Human & $96.24\pm0.67$ & $94.57\pm1.10$ \\ \midrule
DR-GAN~\cite{tran2017disentangled} & $97.84\pm0.79$ & $93.41\pm1.17$ \\
DR-GAN$_{\text{split+inter}}$ & $98.13\pm0.81$ & $93.64\pm1.51$ \\ 
\DrGanAM & $98.36\pm0.75$ & $\mathbf{93.89}\pm1.39$ \\ 
\bottomrule
\end{tabular}
\end{center}
\figvspace
\end{table}

\begin{table}[t!]
\caption{\small Identification rate ($\%$) comparison on Multi-PIE dataset.}
\label{tab:MTPIE_results} \vspace{-3mm}
\begin{center}
\small
\begin{tabular}{ lccccc@{\hskip 1.5mm}c }
\toprule
Method & $0^{\circ}$ & $15^{\circ}$ & $30^{\circ}$ & $45^{\circ}$ & $60^{\circ}$ &  Average\\ \midrule
Zhu et al.~\cite{zhu2013deep} & $94.3$ & $90.7$ & $80.7$ & $64.1$ & $45.9$  & $72.9$ \\
Zhu et al.~\cite{zhu2014multi} & $95.7$ & $92.8$ & $83.7$ & $72.9$ & $60.1$  & $79.3$ \\
Yim et al.~\cite{yim2015rotating} & $\mathbf{99.5}$ & $\mathbf{95.0}$ & $88.5$ & $79.9$ & $61.9$ & $83.3$ \\ 
Using $L2$ loss & $95.1$ & $90.8$ & $82.7$ & $72.7$ & $57.9$  & $78.3$ \\
\midrule
DR-GAN~\cite{tran2017disentangled} & $97.0$ & $94.0$ & $90.1$ & $86.2$ & $83.2$ & $89.2$ \\
DR-GAN & $98.1$ & $94.9$ & $91.1$ & $87.2$ & $84.6$ & $90.4$ \\
\DrGanAM & $98.1$ & $\mathbf{95.0}$ & $\mathbf{91.3}$ & $\mathbf{88.0}$ & $\mathbf{85.8}$ & $\mathbf{90.8}$ \\
\bottomrule
\end{tabular}
\end{center}\figvspace
\end{table}

Table~\ref{tab:CFP_results} shows the comparison on CFP evaluated with Accuracy. %
Results are reported with the average with standard deviation over $10$ folds. 
Overall, we achieve comparable performance on frontal-frontal verification while having $1.92\%$ improvement on the frontal-profile verification. 

Table~\ref{tab:MTPIE_results} shows the face identification performance on Multi-PIE compared to the methods with the same setting. 
Our method shows a significant improvement for large-pose faces, e.g., there is more than $20\%$ improvement margin at $\pm 60^\circ$ poses.
The variation of recognition rates across different poses is much smaller than the baselines, which suggests that our learnt representation is more robust to the pose variation.

\Paragraph{Representation vs.~Synthetic Image for PIFR}
Many prior work~\cite{hassner2015effective, zhu2015high} use frontalized faces for PIFR. 
To evaluate the identity preservation of synthetic images from DR-GAN, we also perform face recognition using our frontalized faces. 
Any face feature extractor could be applied to them, including $G_{enc}$ or $D^d$.
However, both are trained on real images of various poses. 
To specialize to synthetic frontal faces, we fine-tune $G_{enc}$ with the synthetic images and denote as $f'(\cdot)$. 
As shown in Tab.~\ref{tab:ijb-a_multi-cue}, although the performance of synthetic images (and its score-level fusion denoted as $f'(\hat{\mathbf{x}}) \& f(\hat{\mathbf{x}})$) is not as good as the learnt representation, using the fine-tuned $G_{enc}$ on synthetic frontal still achieves comparable perfromance to the previous methods, which shows the identity preservation ability of DR-GAN.

\begin{table}[t!]
\caption{\small{Representation $f(\mathbf{x})$  vs.~synthetic image $\hat{\mathbf{x}}$ on IJB-A.}}
\vspace{-3mm}
\small
\begin{center}
\begin{tabular}{@{\hskip .5mm}c@{\hskip 1.5mm}c@{\hskip 1.5mm}c@{\hskip 1.5mm}c@{\hskip 1.5mm}c@{\hskip .5mm}}
\toprule
& \multicolumn{2}{c}{Verification} & \multicolumn{2}{c}{Identification} \\ \cmidrule(r){2-3} \cmidrule(r){4-5}
Features & @FAR=$.01$ & @FAR=$.001$ & @Rank-$1$ & @Rank-$5$ \\ \midrule
$f(\hat{\mathbf{x}})$ &  $78.5 \pm 1.9$ & $60.3 \pm 3.7$ & $86.9\pm1.6$ & $94.2\pm1.3$   \\
$D^d(\hat{\mathbf{x}})$ & $77.1 \pm 2.9$ & $53.5 \pm 6.2$ & $85.7\pm1.7$ & $93.6\pm1.6$\\
$f'(\hat{\mathbf{x}})$ & $79.2 \pm 2.9$ & $60.8 \pm 7.3$ & $89.2\pm1.4$ & $95.3\pm1.1$\\
$f'(\hat{\mathbf{x}}) \& f(\hat{\mathbf{x}})$ & $83.0 \pm 1.8$ & $71.7 \pm 3.6$ & $90.7\pm1.4$ & ${\bf{95.6}}\pm1.0$\\
$f(\mathbf{x})$ & $\mathbf{84.3}\pm1.4$ & $\mathbf{72.6}\pm4.4$ & $\mathbf{91.0}\pm1.5$ & $\mathbf{95.6}\pm1.1$ \\ \bottomrule
\end{tabular}
\end{center}
\eqnvspace
\label{tab:ijb-a_multi-cue}
\vspace{-2mm}
\end{table}

\begin{figure*}[t!]
\begin{center}
\includegraphics[width=0.85\textwidth]{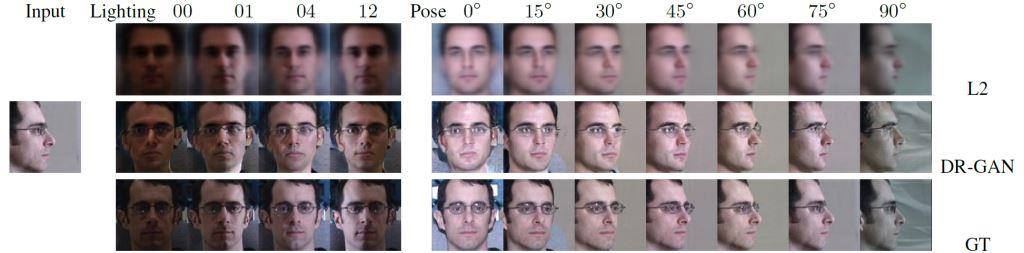}
\vspace{-2mm}
\caption{\small Face rotation comparison on Multi-PIE. Given the input (in illumination $07$ and $75^\circ$ pose), we show synthetic images of $L2$ loss (top), adversarial loss (middle), and ground truth (bottom). Column $2$-$5$ show the ability of DR-GAN in simultaneous face rotation and re-lighting.} \vspace{-1mm}
\label{fig:GANvsL2}
\end{center}
\end{figure*}
\vspace{-2mm}

\begin{figure*}[t!]
  \begin{center}
  \small
\includegraphics[width=0.87\textwidth]{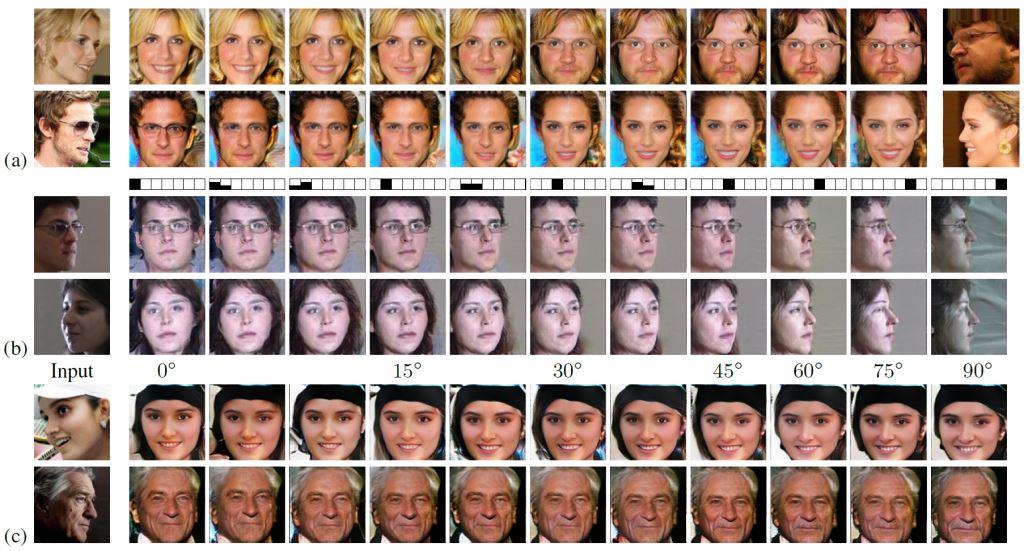} 
\vspace{-1mm}
\caption{\small Interpolation of $f(\mathbf{x})$, $\mathbf{c}$, and $\mathbf{z}$. (a) Synthetic images by interpolating between the identity representations of two faces (Column $1$ and $12$). Note the smooth transition between different genders and facial attributes. (b) Pose angles $0^\circ, 15^\circ, 30^\circ, 45^\circ, 60^\circ, 75^\circ, 90^\circ$ are available in the training set. DR-GAN interpolates in-between {\it unseen} poses via {\it continuous} pose codes, shown above Row $3$. (c) For each image at Column $1$, DR-GAN synthesizes two images at $\mathbf{z}=\mathbf{-1}$ (Column $2$) and $\mathbf{z}=\mathbf{1}$ (Column $12$), and in-between images by interpolating along two $\mathbf{z}$. 
} 
\label{fig:interpolation}
\figvspace \vspace{-1mm}
\end{center}
\end{figure*}

\begin{figure*}[t!]
\begin{center}
\includegraphics[width=0.87\textwidth]{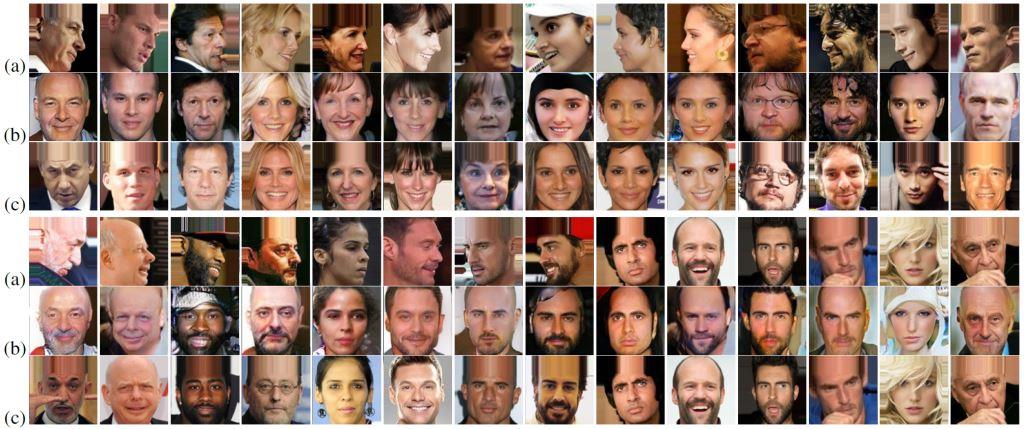}
\end{center}
\vspace{-3mm}
\caption{Face rotation on CFP: (a) input, (b) frontalized faces, (c) real frontal faces, (d) rotated faces at $15^\circ$, $30^\circ$, $45^\circ$ poses.
We expect the frontalized faces to preserve the identity, rather than all facial attributes. 
This is very challenging for face rotation due to the in-the-wild variations and extreme profile views. The artifact in the image boundary is due to image extrapolation in pre-processing. When the inputs are frontal faces with variations in roll, expression, or occlusions, the synthetic faces can remove these variations.}
\label{fig:CFP_Visual}
\end{figure*}

\begin{figure*}[t!]
\begin{center}
\small
\includegraphics[width=0.98\textwidth]{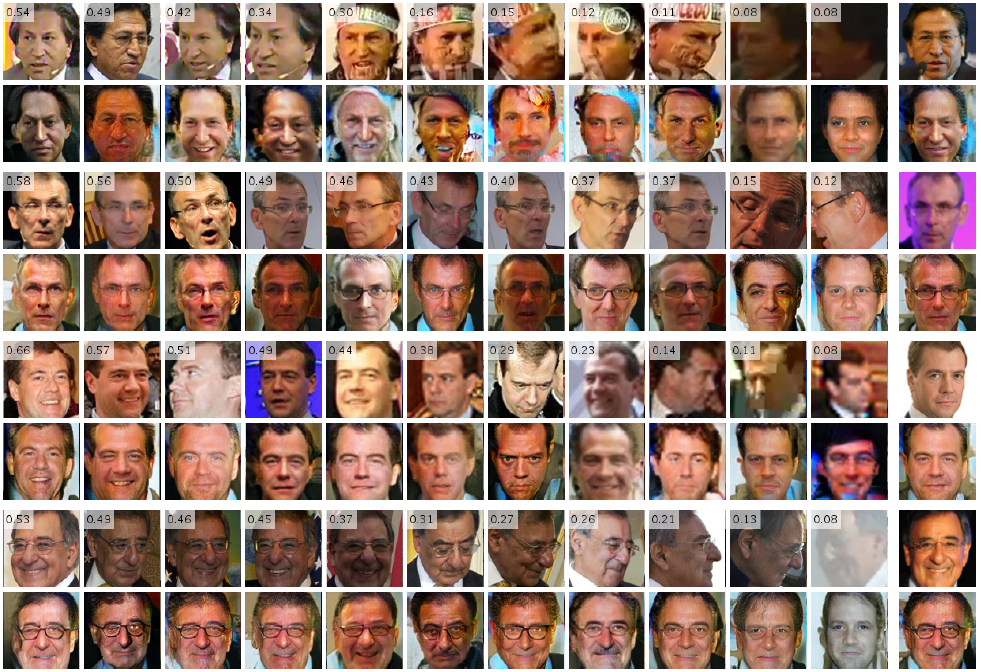}
\figvspace
\caption{\small Face frontalization on IJB-A. 
For each of four subjects, we show $11$ input images with estimated coefficients overlaid at the top left corner (first row) and their frontalized counter part (second row). The last column is the groundtruth frontal and synthetic frontal from the fused representation of all $11$ images. Note the challenges of large poses, occlusion, and low resolution, and our {\it opportunistic} frontalization.}
\label{fig:IJBA_Visual}\figvspace
\end{center}
\end{figure*}

\begin{figure*}[t!]
\begin{center}
\small
\includegraphics[width=0.98\textwidth]{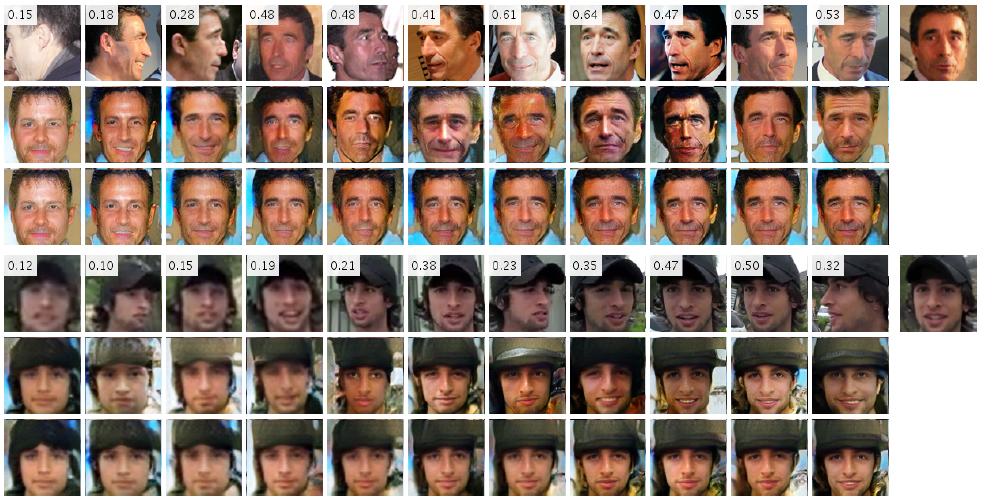}
\vspace{-2mm}
\caption{\small Face frontalization on IJB-A for an image set (first subject) and a video sequence (second subject). 
For each subject, we show $11$ input images (first row), their respective frontalized faces (second row) and the frontalized faces using {\it incrementally} fused representations from all previous inputs up to this image (third row). In the last column, we show the groundtruth frontal face.}
\label{fig:IJBA_num}\figvspace
\end{center}
\end{figure*}

\subsection{Face Rotation}
\Paragraph{Adversarial Loss vs. L2 loss}
Prior work~\cite{zhu2013deep, yim2015rotating, yang2015weakly} on face rotation normally employ the $L2$ loss to learn a mapping between two views.
To compare the $L2$ loss with our adversarial loss, we train a model where $G$ is supervised by an $L2$ loss on the ground truth face with the target view.
The training process is kept the same for a fair comparison. 
As shown in Fig.~\ref{fig:GANvsL2}, DR-GAN can generate far more realistic faces that are similar to the ground truth faces in all views. 
Meanwhile, images synthesized by the $L2$ loss cannot maintain high frequency components and are blurry. 
In fact, $L2$ loss treats each pixel equally, which leads to the loss of discriminative information. 
This inferior synthesis is also reflected in the lower PIFR performance in Tab.~\ref{tab:MTPIE_results}. 
In contrast, by integrating the adversarial loss, we expect to learn a more discriminative representation for better recognition, and a more generative representation for better face synthesis.

\Paragraph{Variable Interpolations}
Taking two images of different subjects $\mathbf{x}_1, \mathbf{x}_2$, we extract features $f(\mathbf{x}_1)$ and $f(\mathbf{x}_2)$ from $G_{enc}$. 
The interpolation between $f(\mathbf{x}_1)$ and $f(\mathbf{x}_2)$ can generate many representations, which can be fed to $G_{dec}$ to synthesize face images. 
In Fig.~\ref{fig:interpolation} (a), the top row shows a transition from a female subject to a male subject with beard and glasses. 
Similar to~\cite{radford2015unsupervised}, these smooth semantic changes indicate that the model has learned essential identity representations for image synthesis. 

Similar interpolation can be conducted for the pose codes as well.
During training, we use a one-hot vector $\bf{c}$ to specify the {\it discrete} pose of the synthetic image. 
During testing, we could generate face images with {\it continuous} poses, whose pose code is the weighted average, i.e., interpolation, of two neighboring pose codes. 
Note that the resultant pose code is no longer a one-hot vector. 
As in Fig.~\ref{fig:interpolation} (b), this leads to smooth pose transition from one view to many views {\it unseen} to the training set.

We can also interpolate the noise vector $\bf{z}$.
We synthesize frontal faces at ${\bf{z}}=-{\bf{1}}$ and ${\bf{z}}={\bf{1}}$ (a vector of all $1$s) and interpolate between two ${\bf{z}}$.
Given the fixed identity representation and pose code, the synthetic images are identity-preserved frontal faces. 
As in Fig.~\ref{fig:interpolation} (c), the change of $\bf{z}$ leads to the change of the background, illumination condition, and facial attributes such as beard, while the identity is well preserved and faces are of the frontal view. 
Thus, $\bf{z}$ models less significant face variations. 

\Paragraph{Face Rotation on Benchmark Databases}
Our generator is trained to be a face rotator.
Given one or multiple face images with arbitrary poses, we can generate multiple identity-preserved faces at different views. 
Figure~\ref{fig:GANvsL2} shows the face rotation results on Multi-PIE. 
Given an input image at any pose, we can generate multi-view images of the same subject but at a different pose by specifying different pose codes or in a different lighting condition by varying illumination code. 
The rotated faces are similar to the ground truth with well-preserved attributes such as eyeglasses. 

One application of face rotation is face frontalization. 
Our DR-GAN can be used for face frontalization by specifying the frontal-view as the target pose. 
Figure~\ref{fig:CFP_Visual} shows the face frontalization on CFP. 
Given an extreme profile input image, DR-GAN can generate a realistic frontal face that has similar identity characteristics as the real frontal face.
To the best of our knowledge, this is the first work that is able to {\it frontalize a profile-view in-the-wild face image}. 
When the input image is already in the frontal view, the synthetic images can correct the pitch and roll angles, normalize illumination and expression, and impute occluded facial areas, as shown in the last few examples of Fig.~\ref{fig:CFP_Visual}.

Figure~\ref{fig:IJBA_Visual} shows face frontalization results on IJB-A. 
For each subject or template, we show $11$ images and their respective frontalized faces, and the frontalized face generated from the fused representation. 
For each input image, the estimated coefficient $\omega$ is shown on the top-left corner of each image, which clearly indicates the quality of the input image as well as the frontalized image. 
For example, coefficients for low-quality or large-pose input images are very small. 
These images will have very little contribution to the fused representation. 
Finally, the face from the fused representation has superior quality compared to all frontalized images from a single input face. 
This shows the effectiveness of our multi-image DR-GAN in taking advantage of multiple images of the same subject for better representation learning. 

To further evaluate face frontalization results w.r.t.~different numbers of input images, we vary the number of input images from $1$ to $11$ and visualize the frontalized images from the {\it incrementally} fused representations. 
As shown in Fig.~\ref{fig:IJBA_num}, the individually frontalized faces have varying degrees of resemblance to the true subject, according to the qualities of different input images. 
The synthetic images from fused representations (third row) improve as the number of images increases. 

%% file: PAMI_2017_DR_GAN_con.tex
\Section{Conclusions}
This paper presents DR-GAN to learn a disentangled representation for PIFR, by modeling the face rotation process. 
We are the first to construct the generator in GAN with an encoder-decoder structure for representation learning, which can be quantitatively evaluated by performing PIFR.  
Using the pose code for decoding and pose classification in the discriminator lead to the disentanglement of pose variation from the identity features. 
We also propose multi-image DR-GAN to leverage multiple images per subject in both training and testing to learn a better representation. 
This is the first work that is able to frontalize an extreme-pose in-the-wild face. 
We attribute the superior PIFR and face synthesis capabilities to the discriminative yet generative representation learned in $G$. 
Our representation is discriminative since the other variations are explicitly disentangled by the pose/illumination codes, and random noise, and is generative since its decoded (synthetic) image would still be classified as the original identity.


%% file: bio/LuanTran.tex
\begin{IEEEbiography}
[{\includegraphics[width=1in,height=1.2in,clip,keepaspectratio]{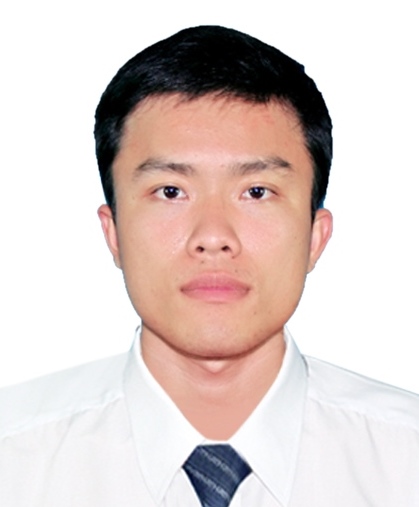}}]
{Luan Tran}
received his B.S. in Computer Science from Michigan State University
with High Hornors in $2015$. He is now pursuing his Ph.D. also at Michigan State University in the area of deep learning and computer vision. His research areas of interest include deep learning and computer vision, in particular, face modeling and face recognition. He is a member of the IEEE.
\end{IEEEbiography}

%% file: bio/XiYin.tex
\begin{IEEEbiography}[{\includegraphics[width=1in,height=1.2in,clip,keepaspectratio]{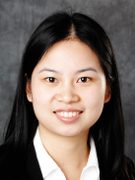}}]{Xi Yin} received the B.S. degree in Electronic and Information Science from Wuhan University, China, in $2013$. Since August $2013$, she has been working toward her Ph.D. degree in the Department of Computer Science and Engineering, Michigan State University, USA. Her research area are face recognition, deep learning, and plant image processing. Her paper on multi-leaf segmentation won the Best Student Paper Award at Winter Conference on Application of Computer Vision (WACV) $2014$. 
\end{IEEEbiography}

%% file: bio/XiaomingLiu.tex
\begin{IEEEbiography}[{\includegraphics[width=1in,height=1.2in,clip,keepaspectratio]{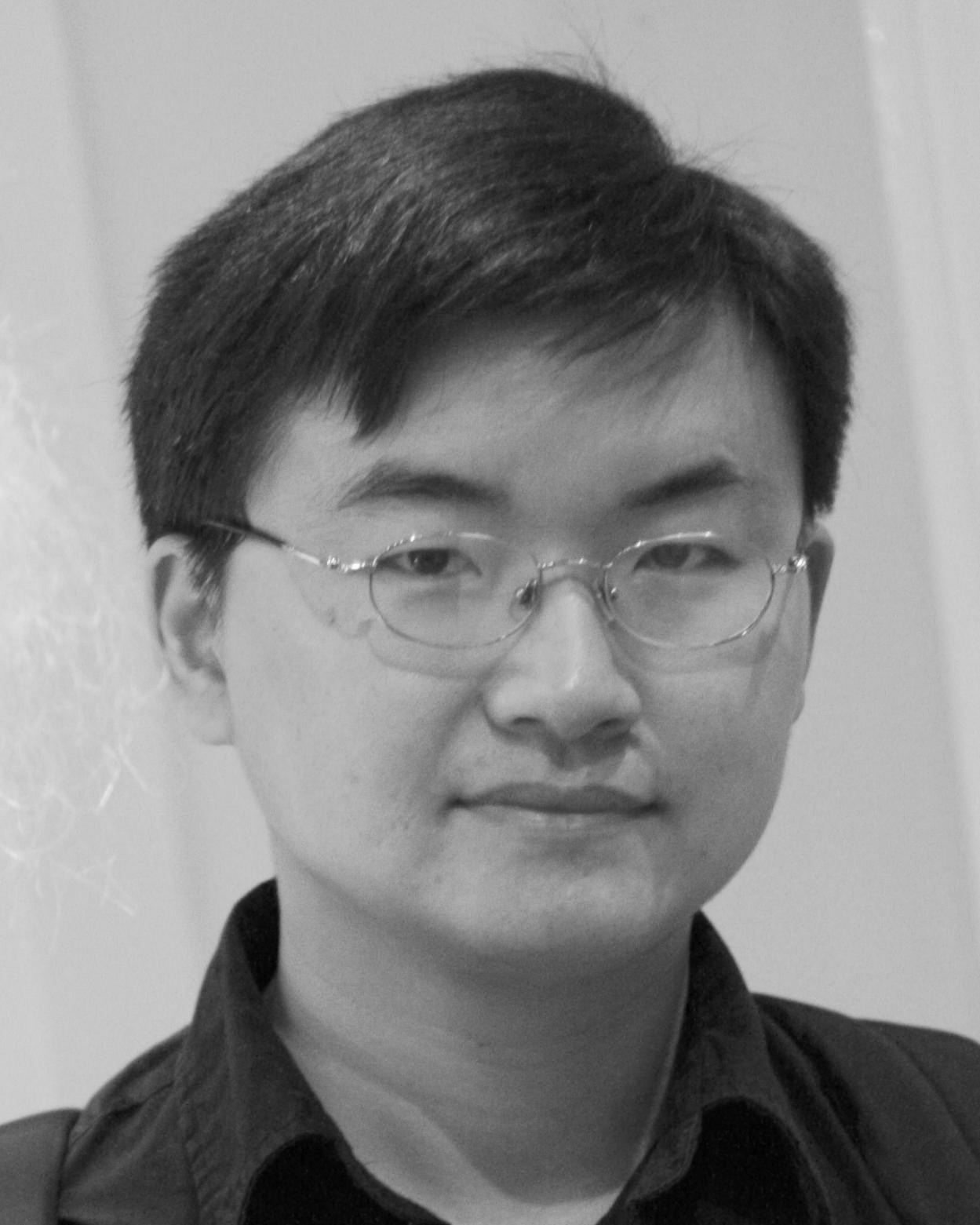}}]{Xiaoming Liu}is an Associate Professor at the Department of Computer Science and Engineering of Michigan State University. He received the Ph.D. degree in Electrical and Computer Engineering from Carnegie Mellon University in $2004$. Before joining MSU in Fall $2012$, he was a research scientist at General Electric (GE) Global Research. His research interests include computer vision, machine learning, and biometrics. As a co-author, he is a recipient of Best Industry Related Paper Award runner-up at ICPR $2014$, Best Student Paper Award at WACV $2012$ and $2014$, and Best Poster Award at BMVC $2015$. He has been the Area Chair for numerous conferences, including FG, ICPR, WACV, ICIP, and CVPR. He is the program co-chair of WACV $2018$ and BTAS $2018$. He is an Associate Editor of Neurocomputing journal. He has authored more than $100$ scientific publications, and has filed $26$ U.S. patents. 
\end{IEEEbiography}